\documentclass[pdflatex,sn-mathphys-num]{sn-jnl}% Math and Physical Sciences Numbered Reference Style
\usepackage{graphicx}%
\usepackage{multirow}%
\usepackage{amsmath,amssymb,amsfonts}%
\usepackage{amsthm}%
\usepackage{mathrsfs}%
\usepackage[title]{appendix}%
\usepackage{xcolor}%
\usepackage{textcomp}%
\usepackage{manyfoot}%
\usepackage{booktabs}%
\usepackage{algorithm}%
\usepackage{algorithmicx}%
\usepackage{algpseudocode}%
\usepackage{listings}%
\usepackage{url}
\usepackage{hyperref}
\usepackage{booktabs}
\usepackage{siunitx}
\usepackage{adjustbox}

\theoremstyle{thmstyleone}%
\theoremstyle{thmstyletwo}%

\theoremstyle{thmstylethree}%

\usepackage{acronym}
\acrodef{FWMAV}[FWMAV]{Flapping-wing Micro Aerial Vehicle}
\acrodef{SWaP}[SWaP]{Size, Weight, and Power}
\acrodef{DEA}[DEA]{Dielectric Elastomer Actuator}
\acrodef{CFD}[CFD]{Computational Fluid Dynamics}
\acrodef{MoI}[MoI]{Moment of Inertia}
\acrodef{IMU}[IMU]{Inertial Measurement Unit}
\acrodef{PWM}[PWM]{Pulse Width Modulation}
\acrodef{CG}[CG]{Center of Gravity}
\acrodef{IIR}[IIR]{Infinite Impulse Response}
\acrodef{CPG}[CPG]{Central Pattern Generator}
\acrodef{PID}[PID]{Proportional-Integral-Derivative}
\acrodef{RLS}[RLS]{Recursive Least Squares}
\acrodef{STAR}[STAR]{Stroke Timing Asymmetry Rhythm}
\acrodef{6-DOF}[6-DOF]{six-degree-of-freedom}
\acrodef{LEV}[LEV]{Leading-Edge Vortex}
\acrodef{OE}[OE]{Output-Error}

\newcommand{\rev}[1]{\textcolor{black}{#1}}

\begin{document}

\title[Article Title]{A 26-Gram Tailless Butterfly-Inspired Flapping-Wing Robot with Onboard Attitude Control}


\author[1]{\fnm{Weibin} \sur{Gu}}

\author[1]{\fnm{Chenrui} \sur{Feng}}%\email{iiauthor@gmail.com}

\author[1]{\fnm{Lian} \sur{Liu}}%\email{iiauthor@gmail.com}

\author[1]{\fnm{Chen} \sur{Yang}}%\email{iiauthor@gmail.com}

\author[1]{\fnm{Xingchi} \sur{Jiao}}%\email{iiauthor@gmail.com}

\author[1]{\fnm{Yuhe} \sur{Ding}}%\email{iiauthor@gmail.com}

\author[1]{\fnm{Xiaofei} \sur{Shi}}%\email{iiauthor@gmail.com}

\author*[1,2]{\fnm{Chao} \sur{Gao}}\email{chao.gao@cantab.net}

\author[3]{\fnm{Alessandro} \sur{Rizzo}}%\email{iiauthor@gmail.com}

\author*[1,4]{\fnm{Guyue} \sur{Zhou}}\email{zhouguyue@air.tsinghua.edu.cn}

\affil*[1]{\orgdiv{Institute for AI Industry Research (AIR)}, \orgname{Tsinghua University}, \orgaddress{\city{Beijing}, \postcode{100084}, \country{PR China}}}

\affil[2]{\orgname{Xinchen Qihang Inc.}, \orgaddress{\city{Beijing}, \postcode{100084}, \country{PR China}}}

\affil[3]{\orgdiv{Department of Electronics and Telecommunications}, \orgname{Politecnico di Torino}, \orgaddress{\city{Turin}, \postcode{10129}, \country{Italy}}}

\affil[4]{\orgdiv{School of Vehicle and Mobility}, \orgname{Tsinghua University}, \orgaddress{\city{Beijing}, \postcode{100084}, \country{PR China}}}


\abstract{
\rev{Butterfly-inspired flapping-wing robots use broad compliant wings and low-frequency actuation, but pronounced wingbeat-synchronous body dynamics and time-varying inertia complicate onboard control. Here, we introduce \textit{AirPulse}, a 26~g two-winged tailless flapping-wing robot with integrated venation-inspired wings, sensing, computation, and power, representing the lowest flight-ready mass among compared butterfly-inspired platforms. We analyze the robot’s dynamic structural coupling and map prescribed modulation parameters of flapping kinematics to experimental six-axis wrench profiles for systematic control channel allocation. To enable stable flapping motion, we formulate Stroke Timing Asymmetry Rhythm (STAR), a phase-domain modulation strategy that ensures smooth stroke velocity changes while preserving mean flapping frequencies. Coupled with state estimation, the onboard feedback control architecture demonstrates successful untethered pitch and directional tracking during climbing and turning maneuvers. Ultimately, the AirPulse robot offers an experimentally validated framework for stabilizing strongly oscillatory, low-mass bio-inspired platforms, providing a basis for future operation in sensitive, confined environments.}

}




\maketitle

\section*{Introduction}\label{sec1}

\acp{FWMAV} have long drawn inspiration from nature to replicate the agility, efficiency, and maneuverability of flying animals~\cite{rafee2025review}. At micro-scales, flapping wings are particularly advantageous as they leverage unsteady lift mechanisms to maintain aerodynamic efficiency where traditional rotary- and fixed-wing designs suffer from severe \rev{low-Reynolds-number} effects. Considerable progress has been achieved at the insect-scale ($<$1~g) using high-bandwidth soft actuators such as piezoelectric bimorphs~\cite{wood2005optimal} and dielectric elastomer actuators~\cite{chen2019controlled}, leading to milestone demonstrations of robotic flies and bees~\cite{wood2008first, yang2019bee+, jafferis2019untethered, ozaki2021wireless, chukewad2021robofly, bena2023high}. However, these sub-gram platforms remain largely tethered or duration-limited due to the absence of power-dense materials capable of replicating insect biomechanics. To overcome this, researchers have designed small-scale \acp{FWMAV} under 100~g~\cite{rafee2025review}, integrating onboard computation and power within strict \ac{SWaP} constraints. These systems, inspired by birds~\cite{yang2012micro,keennon2012development,zhang2017design,yang2018dove,tu2020scale,huang2022all,phan2024twist}, bats~\cite{ramezani2016bat,ramezani2017biomimetic,hoff2021bat,sihite2021integrated}, beetles~\cite{phan2017design,phan2020towards,phan2024passive}, dragonflies~\cite{gaissert2013inventing}, fruit flies and more~\cite{de2009design,richter2011untethered,nguyen2015performance,de2016delfly,roshanbin2017colibri,kiani2019new,karasek2018tailless,de2018quad,chin2020efficient,wu2024multi}, have \rev{consequently supported longer and more complex flight demonstrations compared with sub-gram insect-scale flapping-wing platforms}, marking an important evolution in flapping-wing robotics (Fig.~\ref{fig:Fig7_Discussion}A).

Amidst these advances, one of nature’s most aerodynamically sophisticated fliers, the butterfly (Papilionoidea), remains surprisingly underexplored. Butterflies employ a unique flight paradigm distinct from both fast-flapping insects and larger vertebrate flyers. Unlike birds or bats that rely on active wing morphing or tail actuation~\cite{harvey2022birds, hedenstrom2007bat, de2016delfly}, or bees and fruit flies that can flap above 150~Hz~\cite{fry2003aerodynamics}, butterflies generate sufficient lift and achieve controlled tailless flight using highly compliant wings driven anteromotorically by forewings and through large-amplitude strokes at merely around 10~Hz~\cite{zhang2021kinematic, nan2018can}. Their low wing aspect ratios ($<5$) result in the highest wing-area-to-body-mass ratios among flying animals~\cite{ellington1984aerodynamics}. \rev{Consequently, while butterflies exploit familiar unsteady mechanisms such as delayed stall~\cite{birch2001spanwise}, wake capture~\cite{birch2003influence}, and clap-and-fling~\cite{chin2016flapping}, their biomechanical realization is fundamentally different from insects with rigid, high-aspect-ratio wings flapping at high frequencies that existing aerodynamic studies mainly focus on~\cite{dickinson1999wing,dudley2002biomechanics,sane2003aerodynamics}. Pronounced fluid--structure coupling and severe wing--body inertial interactions drive their characteristic body undulation~\cite{fei2016importance,sridhar2019beneficial,sridhar2020geometric}. This renders them inherently underactuated in \ac{6-DOF}, posing a long-standing challenge in understanding how these tailless insects execute maneuvers such as turning~\cite{henningsson2021downstroke}.}

\rev{Deciphering these mechanisms is hindered by the limitations of current analytical approaches.} While high-speed biological videography has documented trajectories for climbing~\cite{kang2018experimental,sridhar2019beneficial}, forward flight~\cite{lin2012significance, zhang2021kinematic}, and turning~\cite{henningsson2021downstroke, fang2024kinematics}, these descriptive studies lack quantitative links to controllable actuation. Conversely, \ac{CFD} simulations provide detailed insight into flow structures~\cite{le2021adaptive,huang2023ustbutterfly}, but typically assume rigid wings and omit both aeroelastic coupling and whole-body dynamics, significantly limiting their biological and engineering relevance. \rev{Constrained by this disconnection between biological observation and dynamic modeling, the translation of lepidopteran flight into engineered systems has been slow~\cite{sato2010development,tanaka2010forward,fujikawa2019development,huang2022development}. Existing platforms, such as Festo’s eMotionButterflies~\cite{festo2015emotionbutterfly} and USTButterfly~\cite{huang2023ustbutterfly}, provide inspiring engineering context by demonstrating untethered flight; however, the former relies on external motion tracking and offboard computation at 32~g, while the latter lacks quantitative analysis and evidence of controllable flight at 107.1~g. Hence, onboard closed-loop control of two-winged tailless \acp{FWMAV} operating at low flapping frequencies with low-aspect-ratio compliant wings remains largely unexplored. Addressing this gap requires synthesizing lightweight hardware, quantitative control-effectiveness mapping, and onboard feedback control within a platform dominated by wingbeat-synchronous body dynamics. In such context, ``butterfly-inspired'' refers strictly to selected morphological abstractions—specifically a swallowtail planform, compliant membrane wings, and venation-informed reinforcement layout—without implying full fidelity to lepidopteran kinematics, aeroelasticity, flow structures, or neuromuscular control.}

% Despite the vast diversity of lepidopteran wing morphologies~\cite{le2019adaptive}, the aerodynamic contributions of these traits remain poorly characterized~\cite{chin2016flapping}. 

\rev{In this work, we present \textit{AirPulse}, a 26~g two-winged tailless flapping-wing robot that integrates butterfly-inspired morphology with onboard sensing, computation, and closed-loop control (Fig.~\ref{fig:Fig1_BiomimeticDesign}A and Supplementary Movie~S1). The primary contribution is the integrated development and experimental validation of onboard attitude control under pronounced wingbeat-synchronous body motion. Among the butterfly-inspired platforms with sustained untethered flight summarized in Table~\ref{tab:butterfly_comparison}, the AirPulse robot has the lowest reported flight-ready mass, which includes the wings, actuators, battery, onboard sensing and computation, fuselage, and optional receiver for safety override (Fig.~\ref{fig:Fig1_BiomimeticDesign}B). The system-level contribution is supported by four interdependent elements. First, the AirPulse robot integrates independently actuated wing assemblies, onboard sensing and computation, and a venation-informed wing structure (Fig.~\ref{fig:Fig1_BiomimeticDesign}G) that imparts non-uniform spatial compliance beyond conventional rod-reinforced or rigid wings commonly used in Diptera- or hummingbird-inspired robots~\cite{zhao2010aerodynamic,guo2025enhancing}. This hardware foundation facilitates the characterization of its moving center of gravity (Fig.~\ref{fig:Fig2_Characteristics}A), time-varying inertia (Fig.~\ref{fig:Fig2_Characteristics}B), and flapping-induced forces and wing deformation (Fig.~\ref{fig:Fig2_Characteristics}, E to K). Second, Stroke Timing Asymmetry Rhythm (STAR) formulates a smooth and stable phase-domain modulation that preserves mean flapping frequency and provides monotonic phase progression with continuous stroke velocity (Fig.~\ref{fig:Fig3_S3M}, E and F). It addresses the kinematic discontinuities and instability issues existing in prior stroke-timing approaches~\cite{xiong2023lift,ma2013controlled,oppenheimer2011dynamics,tu2020scale}. Third, systematic modulation-to-wrench mapping quantifies six-axis responses across flapping amplitude, frequency, stroke offsets, and STAR modulation (Fig.~\ref{fig:Fig4_FT}), establishing control channel selection, command signs, and cross-axis coupling for tailless systems lacking auxiliary surfaces—a characterization unaddressed in prior studies~\cite{festo2015emotionbutterfly,huang2022development,huang2023ustbutterfly,xiong2023lift,huang2024aerodynamic}. Fourth, established Madgwick filtering, \ac{PID} control, and \ac{RLS} estimation serve as enabling methods, utilizing the experimentally verified control channels in the onboard feedback control design. Untethered pitch and directional tracking for climbing and turning maneuvers validate the integration of these elements (Fig.~\ref{fig:Fig5_Control}, D and E, and Supplementary Movie~S4 and S5).  Collectively, these results demonstrate onboard controlled flight in a 26~g platform and establish an experimentally grounded control framework for underactuated flapping-wing systems with pronounced wingbeat-synchronous dynamics.}

\section*{Results}\label{sec2}

\subsection*{\rev{Design of a 26-gram butterfly-inspired robot}}

\begin{figure}[H]
	\centering
	\includegraphics[width=\textwidth]{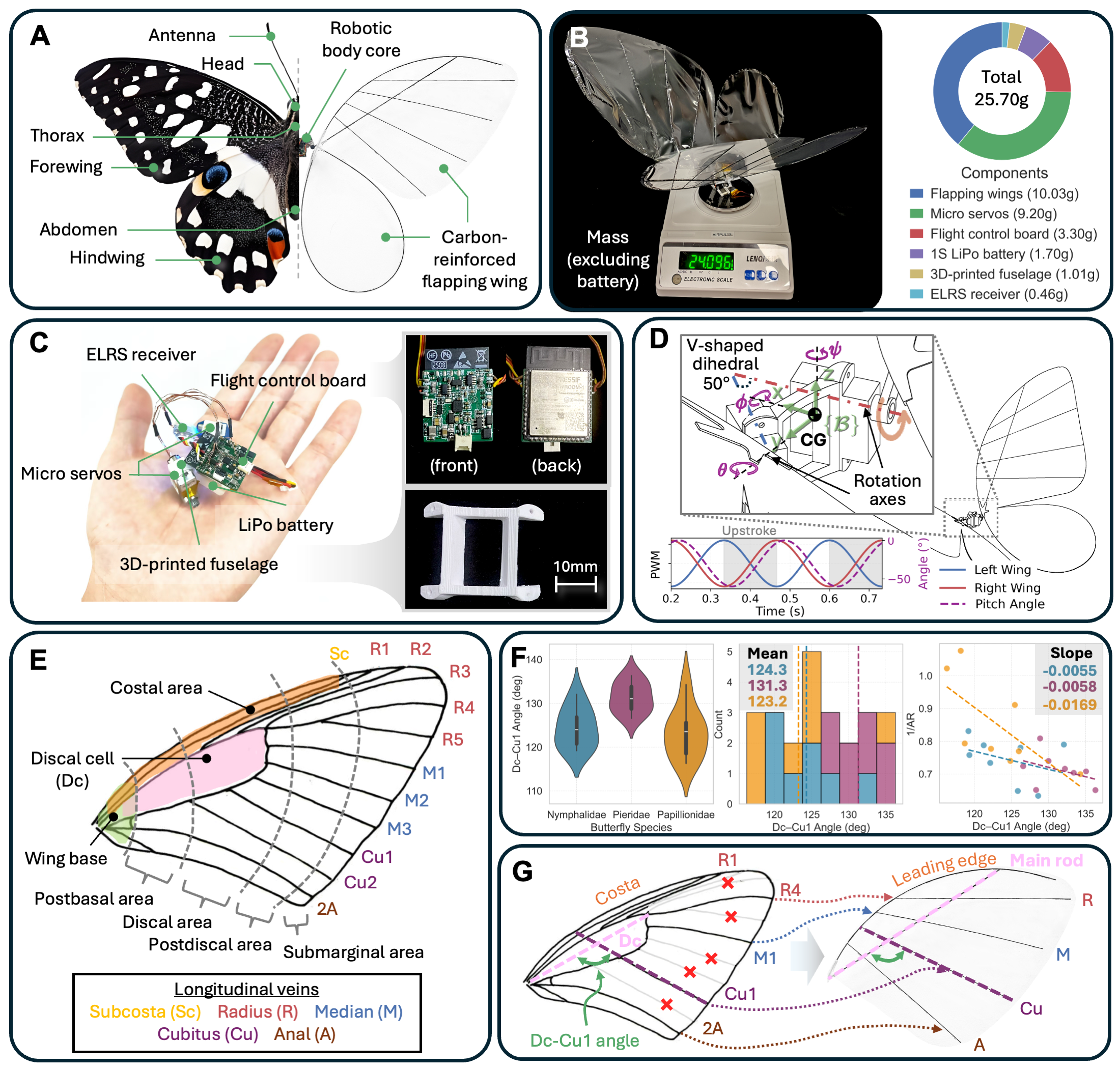} 
    \caption{\rev{\textbf{Morphological and structural design of the AirPulse robot.}
(\textbf{A}) \textit{Papilio demoleus} specimen (left) and AirPulse (right).
(\textbf{B}) Mass distribution of the flight-ready robot.
(\textbf{C}) Body core containing two micro servos, onboard sensing and computation, power electronics, and an ELRS receiver used for safety override. Scale bar, 10~mm.
(\textbf{D}) Actuation schematic and body frame $\mathcal{B}$ ($+x$ forward, $+y$ left, $+z$ up); each mechanically coupled forewing--hindwing assembly is driven by one servo at a fixed dihedral.
(\textbf{E}) Forewing venation of \textit{P. demoleus}, showing the discal cell (Dc) and the subcostal (Sc), radial (R), medial (M), cubital (Cu), and anal (A) vein groups under the Comstock--Needham system~\cite{comstock1898wings}.
(\textbf{F}) Dc--Cu1 angle and geometric slenderness ratio across the surveyed Nymphalidae, Pieridae, and Papilionidae specimens. The slenderness ratio is hind-margin length divided by the corresponding perpendicular width and is distinct from aerodynamic aspect ratio.
(\textbf{G}) Venation-informed carbon-fiber reinforcement layout used for the AirPulse forewing.}}
	\label{fig:Fig1_BiomimeticDesign} 
\end{figure}

\rev{The AirPulse robot is a two-winged tailless flapping-wing robot whose wing planform and reinforcement layout use the lime swallowtail (\textit{Papilio demoleus}) as a morphological reference (Fig.~\ref{fig:Fig1_BiomimeticDesign}A). Its flight-ready mass is approximately 26~g, including two wings, two actuators, battery, flight control unit, fuselage, and safety receiver; the dry mass without the battery is approximately 24~g (Fig.~\ref{fig:Fig1_BiomimeticDesign}B). Among the butterfly-inspired platforms with sustained untethered flight summarized in Table~\ref{tab:butterfly_comparison}, this is the lowest reported flight-ready mass. Festo's eMotionButterflies remain an important prior demonstration of advanced untethered flight using external tracking and offboard control~\cite{festo2015emotionbutterfly}. The distinction of the AirPulse robot is its onboard execution of state estimation, rhythmic-command generation, and closed-loop attitude tracking at this mass. While incorporating butterfly-inspired morphological features, the platform functions as an instrumented robotic system engineered to systematically investigate compliant-wing response, modulation-to-wrench relationships, and onboard feedback control, rather than as a direct biomechanical or aerodynamic replica of a butterfly.}

\rev{The platform was developed through an objective-driven, constraint-based process rather than numerical global optimization. Reducing flight-ready mass decreases inertial loads, actuator demands, and potential impact energy, facilitating safe operation near humans, wildlife, and fragile surroundings. Accordingly, our design objective was to minimize total mass while retaining sufficient force capacity, pitch and directional control authority, structural durability, and the selected compliant butterfly-inspired morphology. The resulting configuration is therefore interpreted as an experimentally validated, constraint-satisfying design within the selected hardware, materials, and manufacturing process, not as a proven optimum in efficiency, controllability, or lift margin. To minimize structural redundancy, the dual micro servos, battery, flight control board, and receiver were integrated directly into a compact 3D-printed fuselage (Fig.~\ref{fig:Fig1_BiomimeticDesign}C)—allowing the \ac{IMU} to be placed close to the \ac{CG}—in contrast to prior designs that mount avionics or batteries on separate, carbon-rod-connected abdominal structures~\cite{festo2015emotionbutterfly,zhang2021kinematic,huang2023ustbutterfly,huang2024aerodynamic}.}

\rev{Wing dimensions were initialized by scaling an approximately 50~g flight-capable prototype (comparable in mass to~\cite{huang2022development}) and then refined experimentally following two stages. The first stage uses geometric and aerodynamic scaling relations~\cite{shyy2013introduction} to initialize the reduction. For an attempted characteristic length ratio, e.g., $\lambda=l_{26}/l_{50}\simeq0.7$, geometric similarity gives $S_{26}/S_{50}\simeq\lambda^2$ and $m_{26}/m_{50}\simeq\lambda^3$, predicting an ideal scaled mass of $50(0.7)^3\simeq17.2~\mathrm{g}$ (where subscripts denote the total configuration weight, $S$ denotes wing area, and $m$ denotes total robot mass). These relations served only as initialization. The approximately $30\%$ wing-to-body mass fraction reported for \textit{Papilio xuthus} specimens~\cite{tanaka2010forward} was likewise used as a mass-allocation reference. Note that although \textit{P. demoleus} provides the primary planform and venation reference in our study, its species-specific wing-mass data were unavailable. Hence, the relative wing mass is compared only with dried-specimen measurements for the congeneric \textit{P. xuthus} with no species-level mechanical equivalence is inferred. In the second stage, wing area, body mass, and wing mass were subsequently reduced while assessing sustained untethered flight performance and actuator reliability. During this two-stage process, we found that strict geometric similarity could not be maintained because the servos, flight control board, and battery were not able to scale continuously with the wing dimensions according to $m\sim l^3$. This limitation establishes what we define as a component-level scaling floor, which requires emerging technologies in actuation and power to achieve further miniaturization. The final AirPulse robot has an approximately 60~cm wingspan, and its wing assemblies account for approximately $38\%$ of the flight-ready mass (Fig.~\ref{fig:Fig1_BiomimeticDesign}, A and B). These values are platform-specific design outcomes rather than universal scaling rules.}

Each lateral wing assembly comprises a forewing and hindwing mechanically coupled via a 3D-printed connector to a common servo arm. Two body-mounted micro servos independently actuate the forewings (Fig.~\ref{fig:Fig1_BiomimeticDesign}D), thereby yielding a coupled, synchronous forewing-hindwing motion with a phase lag (Supplementary Movie~S2). For efficient actuation, the servo axes are canted relative to the longitudinal symmetry plane to align the torque transmission with the structural stiff axis. This configuration produces a fixed $50^{\circ}$ V-shaped dihedral, realized through the fuselage and servo mounts, which alters the mapping of wing-generated forces into body-frame components~\cite{huang2023ustbutterfly}. The broad moving wings produce pronounced wingbeat-synchronous body motion with a phase offset between flapping and body pitch angle (Fig.~\ref{fig:Fig1_BiomimeticDesign}D and Fig.~\ref{fig:Fig5_Control}, F and H). Due to the larger wing-to-body ratio of the robot, this body undulation is more significant than in larger robotic fliers (bird-inspired drones), motivating the control treatment developed later. Besides micro servos, the body core houses a custom flight control board featuring a 9-axis \ac{IMU} and barometer for state estimation, direct \ac{PWM} servo control, and per-rail current monitoring (Fig.~\ref{fig:Fig1_BiomimeticDesign}C and Methods). State estimation, rhythm generation, and feedback control run onboard at 100~Hz. All states are telemetered to the custom ground station for data logging. The \ac{IMU} is mounted near the \ac{CG} and aligned with the body frame ($+x$ forward, $+y$ left, $+z$ up). A power-management circuit supports 1S and 2S LiPo batteries, trading flight endurance against mass.

\rev{Reducing wing dimensions lowers structural mass but also reduces aerodynamic area and lift production available for flapping flight. The principal biologically informed design hypothesis was therefore not to reproduce every feature of a butterfly wing, but to use a sparse, spatially nonuniform reinforcement layout to satisfy two competing requirements: transmitting actuator loads and maintaining the wing planform near the root and leading edge while allowing passive deformation of the more lightly supported membrane. Previous studies have shown that insect-wing venation influences flexural anisotropy and spatial compliance, and that reinforcement topology and passive deformation can alter the aerodynamic response of flapping wings ~\cite{zhao2010aerodynamic,tanaka2010forward,filc2023tailoring}. Motivated by this structural principle,} each wing was fabricated from a $12.5~\mu$m polyethylene terephthalate membrane reinforced with carbon-fiber rods. The primary spanwise spar and leading-edge reinforcement provide the principal load paths between the servo attachment and the distal wing, whereas the intervening and distal membrane regions remain compliant and deform passively during flapping (Supplementary Movie~S2). This venation-informed construction is the principal functional feature transferred from butterfly morphology. It removes the need for an independently actuated wing-pitch degree of freedom, although the present study does not isolate or quantify an energetic benefit of passive feathering.

\rev{The swallowtail-derived planform, broad low-aspect-ratio geometry, and forewing venation pattern serve as secondary morphological references. We surveyed 24 butterfly specimens from Nymphalidae, Pieridae, and Papilionidae to define a reproducible reinforcement geometry (Methods). The mean Dc--Cu1 angles were $131.3^{\circ}\pm3.5^{\circ}$ for Pieridae, $124.3^{\circ}\pm4.6^{\circ}$ for Nymphalidae, and $123.2^{\circ}\pm5.8^{\circ}$ for Papilionidae (Fig.~\ref{fig:Fig1_BiomimeticDesign}F and Supplementary Fig.~\ref{fig:SM_WingStudies}). The angle showed a moderate association with a geometric slenderness ratio defined as hind-margin length divided by the corresponding perpendicular width ($R^2=0.40$, $p=0.0009$). This relationship is descriptive and is not treated as a causal aerodynamic or structural rule. The \textit{P. demoleus} forewing (Fig.~\ref{fig:Fig1_BiomimeticDesign}E) and its approximately $123^{\circ}$ Dc--Cu1 angle were used as the species-specific morphological reference to position simplified R, M, Cu, and A reinforcement analogs around the main spanwise spar for the engineered wing design of the AirPulse robot (Fig.~\ref{fig:Fig1_BiomimeticDesign}G), which contrasts with the venation designs that typically rely on triangular reinforcement structure in the prior butterfly-inspired robots~\cite{festo2015emotionbutterfly, huang2022development,huang2023ustbutterfly}. The complete biological venation pattern was not reproduced because the additional rods required for geometric fidelity conflicted with the robot mass and actuator-load constraints. Each primary longitudinal vein (R, M, Cu, and A) was reinforced by a single carbon rod, aligning with natural venation reduction in lepidopteran lineages such as \textit{Delias eucharis} (e.g., branch loss of R2 and Cu1). The resulting layout is therefore interpreted as a venation-informed engineering abstraction that retains selected major reinforcement directions rather than as a mechanical replica of a butterfly wing. Moreover, these longitudinal rods intersect a main spanwise spar extending from wing base to tip, forming a discal-cell-like structure. The leading edge rod, primary spar, and terminal vein junctions form a costal analog that stiffens the anteromotoric region for efficient torque transfer. Together, these components create a spatially nonuniform reinforcement structure: more heavily reinforced costal and basal regions, a moderately flexible discal zone, and more lightly reinforced distal regions (postdiscal and submarginal) for passive deformation. The hindwing uses a simpler layout without an explicit venation model as a mass--constrained choice; biological evidence concerning hindwing function is used only as qualitative context~\cite{jantzen2008}. Closed periodic cubic B-splines with chord-length parameterization provide reproducible left--right planform geometry (Supplementary Fig.~\ref{fig:SM_WingContourParameterization}; Methods). The present study does not establish that the Dc--Cu1 angle, low aspect ratio, planform, or simplified venation is individually critical or optimal; isolating their effects would require mass- and inertia-matched structural comparisons with direct stiffness measurements.}

\begin{sidewaystable}
    \centering
    \scriptsize
    \caption{\textbf{Comparison of butterfly-inspired flapping-wing robots \rev{with sustained free flight}.}
    \rev{The table provides morphological and flight-ready mass context for the cited platforms and the AirPulse robot.} Wing area $S$ refers to one lateral forewing--hindwing assembly; wing loading and aerodynamic aspect ratio are calculated as $WL=mg/(2S)$ and $AR=b^2/(2S)$, respectively. Only systems with reported sustained free flight are included; prototypes built purely for benchtop characterization or lacking sustained flight capabilities such as those using elastic linkages or rubber-band-driven mechanisms are omitted (e.g.,~\cite{tanaka2010forward,sato2010development,fujikawa2019development,xiong2023lift}).}
    \label{tab:butterfly_comparison}
    \begin{tabular}{lcccc}
        \toprule
        Robot & eMotionButterfly~\cite{festo2015emotionbutterfly} & USTButterfly-II~\cite{huang2022development} & USTButterfly~\cite{huang2023ustbutterfly} & AirPulse (Ours) \\
        \midrule
        Weight ($m$, g) & 32 & 54 & 107.1 & 26 \\
        Wingspan ($b$, cm) & 50 & 50 & 90 & 60 \\
        Aspect Ratio ($\textrm{AR}$) & 2.73$^*$ & 1.76$^*$ & \rev{3.29$^\dag$} & 3.28 \\
        Wing Area ($S$, mm$^2$) & 45\,861.9$^{**}$ & 71\,052.6$^*$ & 123\,184.3 & 54\,916.8 \\
        Wing Loading ($\textrm{WL}$, N\,m$^{-2}$) & \rev{3.42$^*$} & \rev{3.73$^*$} & 4.26$^*$ & 2.32 \\
        \bottomrule
    \end{tabular}
    \vspace{4pt}
    \begin{minipage}{\linewidth}
        \raggedright
        \footnotesize
        $^*$Calculated from available morphological data \rev{using $g=9.81~\mathrm{m\,s^{-2}}$}. \\
        $^{**}$Estimated from geometry using published images. \\
        $^\dagger$\rev{Calculated value from the listed dimensions; reported as 3.81 in~\cite{huang2023ustbutterfly}.}
    \end{minipage}
\end{sidewaystable}

\subsection*{\rev{Wing--body inertial coupling and compliant-wing characterization}}

\begin{figure}[H]
	\centering
	\includegraphics[width=\textwidth]{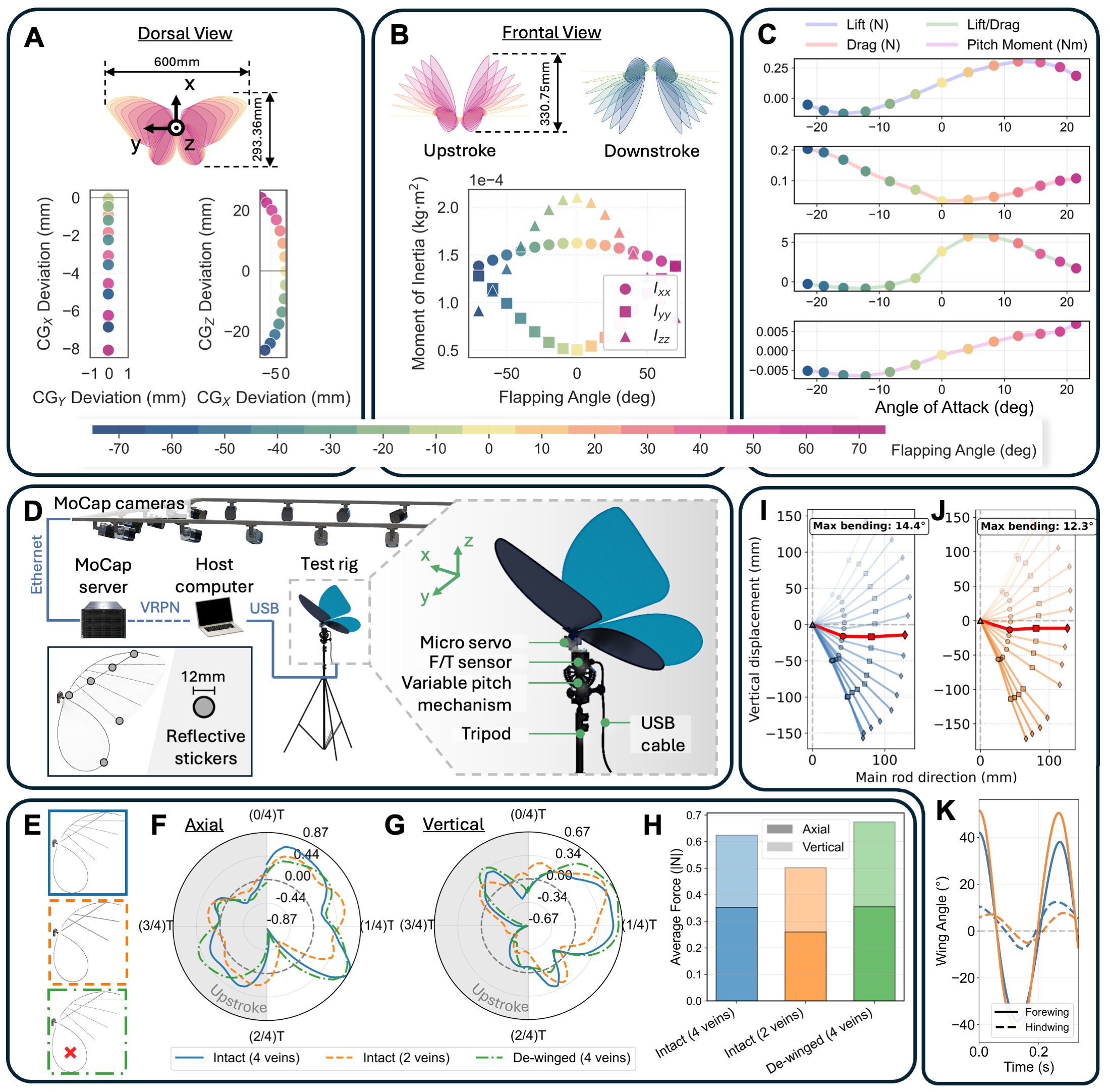} 
\caption{\textbf{Inertial properties and \rev{comparative characterization of the engineered wing configurations}.}
(\textbf{A}, \textbf{B}) Fore--aft and vertical \ac{CG} displacement and \rev{body-axis inertia variation} over one wingstroke.
(\textbf{C}) Static forces and moments measured under a $3~\mathrm{m,s^{-1}}$ horizontal inflow. \rev{Pitching moment is reported using the aerodynamic convention $M_{\mathrm{pitch}}= -M_y^{\mathcal B}$, such that positive values denote nose-up; $M_y^{\mathcal B}$ follows the right-hand-rule convention of the forward--left--up body frame.}
(\textbf{D}) Test rig incorporating a six-axis force--torque (F/T) sensor and a motion-capture system for compliant-wing characterization.
(\textbf{E}) Three tested wing configurations: four-vein full wing (blue), two-vein full wing (orange), and four-vein forewing-only configuration (green).
(\textbf{F}, \textbf{G}) Phase-resolved estimated axial and vertical
aerodynamic forces for the three wing configurations.
(\textbf{H}) Corresponding cycle-averaged axial and vertical forces. 
(\textbf{I}, \textbf{J}) Measured forewing bending angles during flapping for the four-vein and two-vein full-wing configurations.
(\textbf{K}) Chordwise deformation showing a larger distal displacement in the two-vein forewing.}
	\label{fig:Fig2_Characteristics} 
\end{figure}

Unlike conventional aerial robots and fast-flapping insects, the AirPulse robot exhibits pronounced \rev{wingbeat-synchronous} body motion reminiscent of \rev{that observed in} biological butterflies (Fig.~\ref{fig:Fig1_BiomimeticDesign}D and Supplementary Movie~S3). \rev{This behavior arises from the periodic redistribution of mass and angular momentum induced by its comparatively massive wings.} The measured fore--aft \ac{CG} position varies by up to approximately $3\%$ of body length over a wingstroke, while its vertical displacement reaches approximately twice the body core height (Fig.~\ref{fig:Fig2_Characteristics}A). Concurrently, the body-frame inertia components about the pitch and yaw axes ($I_{yy}$ and $I_{zz}$) vary by as much as approximately $2.5\times$ within one wingstroke (Fig.~\ref{fig:Fig2_Characteristics}B).

\rev{To formalize this wing–body coupling, we define an inertial frame $\mathcal{I}$ and a fuselage-fixed body frame $\mathcal{B}$. The total angular momentum about the instantaneous CG (denoted by $G$), ${}^\mathcal{B}\mathbf{H}_G$, can be decomposed as
\begin{equation}
    {}^\mathcal{B}\mathbf{H}_G = \mathbf{I}_G^\mathcal{B}(t) {}^\mathcal{B}\boldsymbol{\omega} + {}^\mathcal{B}\mathbf{h}_{\mathrm{rel}}(t),
\end{equation}
where ${}^\mathcal{B}\boldsymbol{\omega} = [p, q, r]^\top$ is the body-fixed angular rate, ${}^\mathcal{B}\mathbf{h}_{\mathrm{rel}}(t)$ is the relative angular momentum of the wings, and $\mathbf{I}_G^\mathcal{B}(t)$ is the instantaneous inertia tensor of the whole robot, explicitly written as
$\mathbf{I}_G^\mathcal{B}(t) = \begin{bmatrix} I_{xx} & I_{xy} & I_{xz} \\ I_{yx} & I_{yy} & I_{yz} \\ I_{zx} & I_{zy} & I_{zz} \end{bmatrix}.$
Because the principal inertia axes vary dynamically during flapping, $\mathbf{I}_G^\mathcal{B}(t)$ is non-diagonal (\rev{Table~\ref{tab:inertia_variation}}). Consequently, the components $I_{yy}$ and $I_{zz}$ introduced earlier represent time-varying, body-frame inertias rather than constant principal moments. The angular-momentum balance about the instantaneous CG, expressed in the rotating body frame, is
\begin{align}
{}^{\mathcal{B}}\boldsymbol{\tau}_{G,\mathrm{ext}}
&=
\left(\frac{d\,{}^{\mathcal{B}}\mathbf{H}_G}{dt}\right)_{\mathcal{B}}
+
{}^{\mathcal{B}}\boldsymbol{\omega}
\times
{}^{\mathcal{B}}\mathbf{H}_G
\nonumber\\
&=
\mathbf{I}^{\mathcal{B}}_G\,{}^{\mathcal{B}}\dot{\boldsymbol{\omega}}
+
\dot{\mathbf{I}}^{\mathcal{B}}_G\,{}^{\mathcal{B}}\boldsymbol{\omega}
+
{}^{\mathcal{B}}\dot{\mathbf{h}}_{\mathrm{rel}}
+
{}^{\mathcal{B}}\boldsymbol{\omega}
\times
\left(
\mathbf{I}^{\mathcal{B}}_G\,{}^{\mathcal{B}}\boldsymbol{\omega}
+
{}^{\mathcal{B}}\mathbf{h}_{\mathrm{rel}}
\right),
\label{eq:full_angular_momentum_balance}
\end{align}
where ${}^{\mathcal{B}}\boldsymbol{\tau}_{G,\mathrm{ext}}$ denotes the net
external torque about the instantaneous CG expressed in $\mathcal{B}$. The exact $y_{\mathcal{B}}$ component is therefore
\begin{align}
{}^{\mathcal{B}}\tau_{G,y,\mathrm{ext}}
={}&
I_{yx}\dot{p}
+
I_{yy}\dot{q}
+
I_{yz}\dot{r}
+
\dot{I}_{yx}p
+
\dot{I}_{yy}q
+
\dot{I}_{yz}r
+
\dot{h}_{\mathrm{rel},y}
\nonumber\\
&+
\left[
{}^{\mathcal{B}}\boldsymbol{\omega}
\times
\left(
\mathbf{I}^{\mathcal{B}}_G\,{}^{\mathcal{B}}\boldsymbol{\omega}
+
{}^{\mathcal{B}}\mathbf{h}_{\mathrm{rel}}
\right)
\right]_y .
\label{eq:exact_pitch_momentum_balance}
\end{align}}

\rev{For the experimentally observed pitch-dominant motion, $|p|,|r|\ll|q|$, the off-axis acceleration, time-varying cross-inertia, and gyroscopic terms are small relative to the retained pitch-axis contributions. The $y_{\mathcal{B}}$ balance can therefore be approximated as
\begin{equation}
{}^{\mathcal{B}}\tau_{G,y,\mathrm{ext}}
\approx
I_{yy}\dot{q}
+
\dot{I}_{yy}q
+
\dot{h}_{\mathrm{rel},y}.
\label{eq:pitch_momentum_approximation}
\end{equation}
Importantly, this reduction does not assume that the body frame coincides with the instantaneous principal-inertia frame; rather, it follows from the experimentally observed pitch-dominant motion. The term $\dot{I}_{yy}q$ represents the contribution of the time-varying body-frame inertia, whereas $\dot{h}_{\mathrm{rel},y}$ represents changes in angular momentum associated with wing motion relative to the fuselage. For a conventional rigid aircraft, $\dot{\mathbf{I}}_G=\mathbf{0}$ and $\mathbf{h}_{\mathrm{rel}}=\mathbf{0}$, so these additional coupling terms vanish and the standard constant-inertia rigid-body equation is recovered.} In the AirPulse robot, however, $I_{yy}(t)$ varies substantially during each wingstroke. During wing retraction (e.g., from $-70^\circ$ to $0^\circ$; Fig.~\ref{fig:Fig2_Characteristics}B), $\dot{I}_{yy}<0$, while simultaneous changes in ${}^{\mathcal{B}}\mathbf{h}_{\mathrm{rel}}$ redistribute angular momentum between the moving wings and fuselage. Consequently, even in the absence of an external aerodynamic pitching moment, the fuselage can undergo an opposing rotational response to conserve the total angular momentum of the system. This internal wing--body coupling, together with the external aerodynamic moments generated during flapping, contributes to the pronounced wingbeat-synchronous body undulation observed in flight.

% \rev{Substituting this decomposition into the angular-momentum balance and isolating the pitch ($y_B$) axis for the observed pitch-dominant motion ($\vert{}p\vert{}, \vert{}r\vert{} \ll \vert{}q\vert{}$) yields the approximation
% \begin{equation}
%     \tau_{G,y,\mathrm{ext}} \approx I_{yy}\dot{q} + \dot{I}_{yy}q + \dot{h}_{\mathrm{rel},y}.
% \end{equation}}Here, $\dot{I}_{yy}q$ represents the inertial coupling between the time-varying body inertia and pitch rotation (where $q \approx \dot{\theta}$). In conventional aircraft, $I_{yy}$ is nearly constant ($\dot{I}_{yy} \approx 0$), and the body responds solely to external torques ($\tau_{G,y,\mathrm{ext}}$). In the AirPulse robot, however, $I_{yy}(t)$ changes rapidly as the wings accelerate and decelerate. When the wings retract (e.g., from $-70^\circ$ to $0^\circ$; Fig.~\ref{fig:Fig2_Characteristics}B), $\dot{I}_{yy} < 0$, contributing to the redistribution of angular momentum between the wings and fuselage. Through this internal redistribution of angular momentum ($\dot{h}_{\mathrm{rel},y}$), wing acceleration induces an opposing fuselage rotation even in the absence of external aerodynamic torque. This alternating exchange of angular momentum over successive strokes contributes to the observed body undulation.

\rev{A scaling analysis further illustrates how the relative wing--inertial contribution can increase when a thin-membrane flapping robot is reduced in size. Let $l$ denote characteristic length, $f$ flapping frequency, and $\rho$ air density. Under geometric scaling, the weight scales as $mg\sim l^3$, while a characteristic aerodynamic force with wing velocity $U_w\sim fl$ and area $S\sim l^2$ scales as $\rho U_w^2S$. Balancing these quantities gives $f\sim l^{-1/2}$ and a characteristic aerodynamic torque $\tau_{\mathrm{aero}}\sim\rho U_w^2Sl\sim l^4$. For thin engineered wings with approximately scale-invariant areal density, $m_w\sim l^2$ and $I_w\sim m_wl^2\sim l^4$. With $\ddot{\phi}\sim f^2\sim l^{-1}$, the corresponding inertial torque scales as $\tau_{\mathrm{inertial}}\sim I_w\ddot{\phi}\sim l^3$, giving $\tau_{\mathrm{inertial}}/\tau_{\mathrm{aero}}\sim l^{-1}$. This relation is conditional on the stated force-balance and thin-wing assumptions; it is not a universal scaling law for all flapping vehicles. For the AirPulse robot, the wing assemblies constitute approximately $38\%$ of flight-ready mass, so their periodic inertial contribution cannot be neglected.}

Static aerodynamic forces and moments (Fig.~\ref{fig:Fig2_Characteristics}C) were measured with the AirPulse robot mounted on a six-axis force--torque sensor (BOTA MiniOne; 150~mN force resolution and 1.8~mNm torque resolution), with the robot \ac{CG} aligned to the sensor center, installed in a custom open-circuit wind tunnel (1.8~m $\times$ 1.2~m $\times$ 1.0~m, turbulence intensity $<7.5\%$, blockage ratio $<5\%$) at a mean wind speed of $3~\mathrm{m\,s^{-1}}$. \rev{While exceeding conventional low-turbulence thresholds, these measurements are used primarily for within-tunnel relative comparisons because all prescribed static configurations were tested under the same inflow condition. For each prescribed static wing configuration, the effective angle of attack ($\alpha$) was calculated from the wing orientation relative to the freestream. For this static aerodynamic analysis (Fig.~\ref{fig:Fig2_Characteristics}C), we report the pitching moment using the conventional aerodynamic sign convention, with positive moment denoting nose-up; accordingly, $M_{\mathrm{pitch}}= -M_y^{\mathcal B}$, where $M_y^{\mathcal B}$ is the moment about the $+y$ axis of our forward--left--up body frame and is positive nose-down by the right-hand rule. Under this convention, the measured positive slope $\partial M_{\mathrm{pitch}}/\partial\alpha>0$ is, under a quasi-steady interpretation of the prescribed static configurations, consistent with an effective neutral point ahead of the \ac{CG}. This indicates a destabilizing static aerodynamic tendency, but does not determine the longitudinal dynamic stability of the freely flapping robot, whose aerodynamic loads, \ac{CG}, and inertia vary within each wingbeat.} Static lift measurements show a peak of 0.292~N at a $40^\circ$ flapping angle, exceeding the body weight of 0.26~N and suggesting a potential for gliding phases. These measurements, however, do not determine force generation or stability during active flapping.

\rev{To examine how simplified reinforcement layout influences the aeroelastic behavior of the broad, low-aspect-ratio wings,} we compared three engineered wing configurations (Fig.~\ref{fig:Fig2_Characteristics}E): (i) a full forewing--hindwing assembly with four longitudinal forewing reinforcements along the R, M, Cu, and A directions; (ii) a comparable full assembly with two longitudinal forewing reinforcements; and (iii) the four-vein configuration with the hindwing membrane removed. \rev{The first comparison changes forewing reinforcement while retaining the complete planform, while the second examines hindwing-membrane removal from the four-vein design, motivated by observations that hindwings primarily enhance maneuverability rather than generate essential lift~\cite{jantzen2008}. These are comparative engineering tests rather than a factorial stiffness study since changing rod number or hindwing membrane simultaneously changes structural compliance, mass distribution, and wing inertia. The measurements therefore cannot isolate a prescribed stiffness gradient or establish a general aerodynamic benefit of biological venation.}

Each configuration was tested on the motion-capture (LUSTER FZMotion) and force--torque rig (Fig.~\ref{fig:Fig2_Characteristics}D), \rev{and aerodynamic contributions were estimated by the inertial-subtraction procedure (Methods).} The four-vein full-wing and forewing-only configurations exhibited similar phase-resolved axial and vertical force profiles, whereas the two-vein full-wing configuration exhibited different profiles implying a different aeroelastic response (Fig.~\ref{fig:Fig2_Characteristics}, F and G). The four-vein full wing also produced larger cycle-averaged axial and vertical forces than the two-vein full wing but lower values than the four-vein forewing-only configuration (Fig.~\ref{fig:Fig2_Characteristics}H). \rev{The corresponding time-averaged nondimensional force coefficients $(\overline{C}_{F_x},\overline{C}_{F_z})$ were $(6.084, 4.780)$, $(4.519, 4.172)$, and $(7.013, 6.025)$, respectively (Methods). The forewing-only configuration produced larger cycle-averaged force coefficients than the four-reinforcement full-wing configuration under the tested conditions. Because hindwing removal simultaneously changes aerodynamic area, mass distribution, inertia, and structural coupling, the present comparison does not identify the mechanism responsible for this difference. Resolving the respective structural and aerodynamic contributions will require matched structural references together with phase-resolved fluid–structure measurements.} Wing deformation measurements further show that four- and two-vein full wings had similar overall stroke amplitudes and maximum measured bending angles of $14.4^{\circ}$ and $12.3^{\circ}$ near the horizontal downstroke position (Fig.~\ref{fig:Fig2_Characteristics}, I and J), while the two-vein forewing showed larger chordwise displacement, confirming higher
compliance in distal regions after reducing the reinforcement (Fig.~\ref{fig:Fig2_Characteristics}K). \rev{Together, these results provide comparative evidence that reinforcement layout influences the coupled structural and aerodynamic response of the engineered wing.} The observed trend aligns qualitatively with prior work demonstrating that trailing-edge flexibility reduces aerodynamic force generation in isotropic wings with simplified venation~\cite{zhao2010aerodynamic}.

% 1. Full 4 Configuration\(F_{x,\text{full4}} / 0.057530\) = \(\frac{0.35}{0.057530} \approx\) 6.0838\(F_{z,\text{full4}} / 0.057530\) = \(\frac{0.275}{0.057530} \approx\) 4.7801
% 2. Full 2 Configuration\(F_{x,\text{full2}} / 0.057530\) = \(\frac{0.26}{0.057530} \approx\) 4.5194\(F_{z,\text{full2}} / 0.057530\) = \(\frac{0.24}{0.057530} \approx\) 4.1717
% 3. Fore Configuration\(F_{x,\text{fore}} / 0.050621\) = \(\frac{0.355}{0.050621} \approx\) 7.0129\(F_{z,\text{fore}} / 0.050621\) = \(\frac{0.305}{0.050621} \approx\) 6.0252

\subsection*{Stroke Timing Asymmetry Rhythm (STAR): a smooth and stable \rev{phase-domain modulation}}

\begin{figure}[H]
	\centering
	\includegraphics[width=\textwidth]{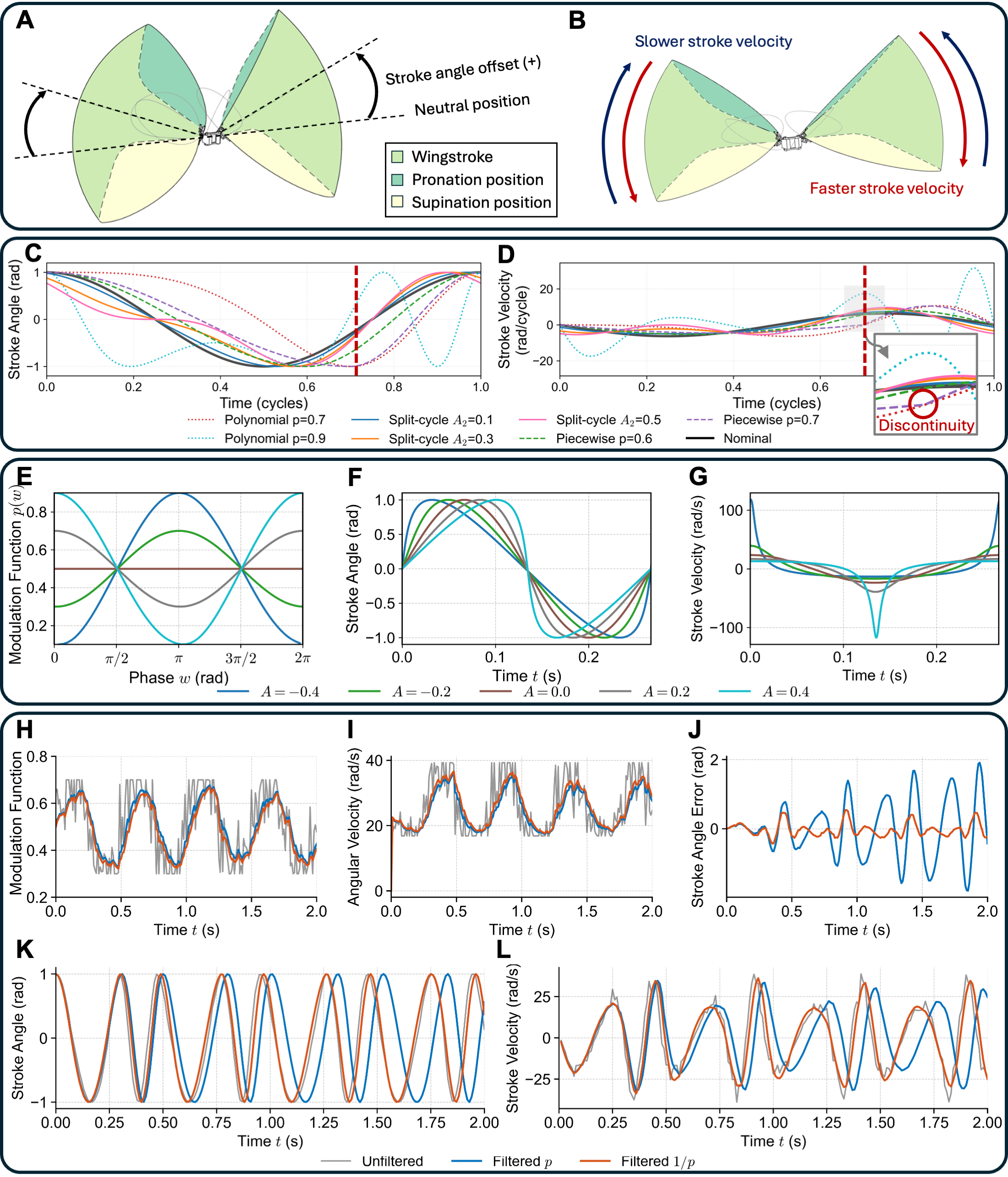} 
\caption{\rev{\textbf{Command-level comparison for flapping modulation and implementation of STAR.}
(\textbf{A}) Stroke-angle-offset modulation changes the neutral stroke position.
(\textbf{B}) Stroke-timing modulation redistributes time and velocity between the two half-strokes while retaining the stroke amplitude.
(\textbf{C}, \textbf{D}) Representative command trajectories generated by previous polynomial phase-shaping and split-cycle formulations, illustrating their parameter-dependent admissibility and smoothness properties.
(\textbf{E}--\textbf{G}) STAR trajectories for different values of $A$. Within $|A|<0.5$, phase progression is monotonic and the stroke angle and velocity remain continuous.
(\textbf{H}, \textbf{I}) Filtering of the reciprocal phase-speed factor $1/p$ during time-varying modulation.
(\textbf{J}--\textbf{L}) STAR with IIR filtering preserves cycle-averaged flapping frequency while enabling stable and linear tuning of stroke asymmetry while reducing phase distortion and cumulative drift relative to filtering $p$ directly.}}
	\label{fig:Fig3_S3M} 
\end{figure}

Flapping-wing attitude control requires low-dimensional modulation of periodic wing motion that remains numerically stable, mechanically realizable, and sufficiently smooth for actuator-limited platforms. In addition to stroke-amplitude and stroke-offset modulation (Fig.~\ref{fig:Fig3_S3M}A)~\cite{phan2020towards,huang2023ustbutterfly, wu2024multi}, the relative durations of the two half-strokes can be varied while preserving the overall stroke amplitude (Fig.~\ref{fig:Fig3_S3M}B). Such timing asymmetry changes the phase-resolved wing velocity and loading and therefore provides an additional control input ~\cite{xiong2023lift,ma2013controlled,oppenheimer2011dynamics,tu2020scale}. \rev{Biological observations provide general motivation for considering wingstroke-timing variation~\cite{fry2003aerodynamics}; however, we shall clarify that the formulation introduced here is an abstract engineering phase modulation rather than a model of insect neuromuscular control.}

Existing stroke-timing laws use different parameterizations and consequently provide different combinations of admissible asymmetry, trajectory smoothness, and period preservation (Fig.~\ref{fig:Fig3_S3M}, C and D). Polynomial phase-shaping can generate differentiable commands but may have a restricted admissible parameter domain~\cite{xiong2023lift}, whereas split-cycle formulations can produce large timing asymmetry but may introduce velocity discontinuities at switching points~\cite{oppenheimer2011dynamics,tu2020scale} or lose viable periodicity~\cite{ma2013controlled} if their parameters are used outside the specified range. These observations motivate a formulation whose admissible domain and kinematic properties can be stated explicitly.

We therefore formulate Stroke Timing Asymmetry Rhythm (STAR), a single-parameter phase-domain modulation that guarantees continuous phase speed and smooth first derivatives, producing bounded, monotonic up–down stroke asymmetry (Fig.~\ref{fig:Fig3_S3M}, E to G). With the phase origin defined at mid-downstroke ($\omega=0$), \ac{STAR} is defined by
\begin{equation} 
\label{eq:S3M} 
y(t)=\zeta\sin\omega(t), \qquad \dot{\omega}(t)=\frac{\pi f}{p(\omega(t))}, \qquad p(\omega)=\frac{1}{2}\rev{-}A\cos\omega, \qquad |A|<\frac{1}{2}, 
\end{equation} 
where $y(t)$ is the commanded stroke trajectory, $\zeta$ is the flapping amplitude, $f$ is the nominal flapping frequency, and $A$ is the dimensionless modulation parameter that controls the timing asymmetry whose admissible range can be linearly normalized to $(0, 1)$ for convenience. Positive $A$ produces a faster downstroke and slower upstroke, whereas negative $A$ reverses this asymmetry. The cosine form of the modulation function $p(\cdot)$ is deliberately chosen to suppress phase offsets arising from initial errors (Supplementary Appendix~\ref{app:star}). Formally, this formulation provides three analytical properties (Methods): (i) the cycle-averaged frequency is exactly preserved for any constant $A$, allowing intra-cycle timing to be skewed without altering mean rate; (ii) stroke asymmetry scales linearly and monotonically with $A$, yielding a simple yet direct tuning rule; (iii) both stroke angle and its first derivative remain continuous, preventing abrupt accelerations that could saturate actuators or excite structural vibrations. 

\rev{The commanded kinematic properties of representative polynomial, split-cycle, and our STAR formulations are compared (Fig.~\ref{fig:Fig3_S3M}, C, D, and F). This comparison is analytical and command-level rather than an experimental hardware benchmark. A controlled energetic comparison would require the methods to produce matched stroke amplitude, mean frequency, half-stroke duration ratio, peak velocity, and peak acceleration. Because the different phase laws cannot in general satisfy all of these conditions simultaneously, differences in measured power would conflate the modulation formulation with differences in the realized kinematics. We therefore do not claim that STAR consumes less power or provides universally better control performance than previous timing methods. The practical contribution of STAR is instead that its admissible parameter domain, period, phase monotonicity, and first-derivative continuity are explicit, and that the resulting commands were implemented on the AirPulse robot and validated through force--torque characterization and untethered closed-loop flight as shown later.}

For time-varying commands, an \ac{IIR} filter is further applied to the reciprocal phase-speed factor $1/p$ rather than the modulation function $p$ itself (Fig.~\ref{fig:Fig3_S3M}, H and I; Methods). This nuance is essential for the time consistency of the rhythm because the instantaneous phase velocity is directly proportional to $1/p$, therefore smoothing this term reduces phase distortion and cumulative drift relative to filtering $p$ directly (Fig.~\ref{fig:Fig3_S3M}, J to L). In contrast, filtering $p$ would introduce a nonlinear lag that distorts the intended stroke timing and induces a cumulative phase drift. By processing the reciprocal variable, we preserve a linear relationship between the modulatory control demands and the resulting wing motion. This ensures that \ac{STAR} remains both responsive to time-varying commands and free from the timing errors that typically plague rhythmic systems, ultimately allowing for more predictable and effective attitude control.

\subsection*{Mapping forces and torques from flapping kinematics}

The flapping motion of butterflies is governed by \acp{CPG}, neural circuits that generate rhythmic motor outputs without the need for rhythmic inputs. These networks exhibit strong adaptability, as their intrinsic neuronal properties and synaptic connections can be dynamically modulated through sensory feedback. For onboard implementation, periodic left- and right-wing commands are generated by an analytical rhythm generator that combines a sinusoidal oscillator with STAR described by
\begin{equation}
\label{eq:STAR}
    y(t) = \zeta \sin(\omega(t)) + \delta,
\end{equation}
where $\delta$ denotes the stroke angle offset and other variables are defined in~\eqref{eq:S3M}. \rev{Notably, we use the term \ac{CPG} in the engineering sense of a rhythmic command generator and~\eqref{eq:STAR} is not a model of butterfly neural circuitry or neuromuscular actuation.} The independently commanded parameters are flapping frequency $f$, amplitude $\zeta$, stroke offset $\delta$, and timing-asymmetry parameter $A$. \rev{Their left--right symmetric and differential combinations provide the candidate control inputs evaluated below.}

% While Hopf oscillators are commonly used for robotic \acp{CPG}, they primarily ensure rhythmic stability during amplitude transitions with fixed phase velocity~\cite{zhang2016instantaneous}. This property contrasts with insect flight, where differential stroke timing rather than amplitude modulation alone serves as the primary control mechanism.

A central challenge in designing flapping-wing robots is to determine control effectiveness, namely, how control inputs or kinematic modulation parameters map to measurable forces and torques for control allocation and dynamical modeling. This challenge is particularly evident in the AirPulse robot, which operates with only two actuators in a \ac{6-DOF} workspace. The resulting underactuation produces coupled and nonlinear force--torque responses across multiple axes. Existing studies of butterfly-inspired robots~\cite{sato2010development,tanaka2010forward,fujikawa2019development,huang2022development,huang2023ustbutterfly} rarely provide systematic quantification of these input--output relationships, leaving the control effectiveness of this class of aerial robots insufficiently characterized. We therefore measured the total flapping-induced forces and torques generated by the AirPulse robot using the test rig (Fig.~\ref{fig:Fig2_Characteristics}D), providing an empirical basis for control allocation and controller design. Although quasi-steady and other reduced-order aerodynamic models can relate prescribed wing kinematics to instantaneous force generation~\cite{ellington1984aerodynamics, sane2003aerodynamics,chin2016flapping}, accurate first-principles prediction is challenging for the present flexible and morphologically heterogeneous wings because the measured response may contain combined contributions from rotational circulation, \ac{LEV} dynamics, wake interaction, added-mass effects, structural deformation, wing inertia, and transmission dynamics. \rev{The present empirical measurements do not resolve or separately identify these mechanisms. Instead, they characterize their aggregate contribution to the force--torque response transmitted through the tested robot and apparatus under prescribed operating conditions. The resulting control-effectiveness mapping is therefore interpreted as system-level input--output characterization.}

Six experimental campaigns systematically varied the key control parameters, including flapping amplitude (Fig.~\ref{fig:Fig4_FT}, A and B), flapping frequency (Fig.~\ref{fig:Fig4_FT}, C and D), symmetric and antisymmetric angle offset modulation (Fig.~\ref{fig:Fig4_FT}, E to G), and symmetric and antisymmetric stroke timing modulation through the \ac{STAR} parameter $A$ (Fig.~\ref{fig:Fig4_FT}, H to J). \rev{For each configuration, total forces and moments were recorded along all three body axes over 80 wingbeats, and the middle 40 wingbeats were phase-aligned and averaged pointwise at each normalized wingbeat phase to obtain a representative phase-averaged force--moment waveform. Each plotted data point in Fig.~\ref{fig:Fig4_FT} represents the temporal mean of this phase-averaged waveform, whereas the error bars denote its standard deviation over wingbeat phase, thereby quantifying the magnitude of periodic intra-wingbeat load variation.}

Increasing either flapping amplitude or frequency increased the cycle-averaged resultant force, predominantly through the body-$x$ component (Fig.~\ref{fig:Fig4_FT}, B and D). \rev{By contrast, the cycle-averaged pitch moment varied by approximately $3.5~\mathrm{mN\,m}$ and $2.1~\mathrm{mN\,m}$ under symmetric amplitude and frequency modulation, respectively, which are substantially less than that produced by stroke-offset and STAR modulation (Supplementary Fig.~\ref{fig:SM_CycleAveragedPitchMoment}). Pitch moment is mainly analyzed because amplitude and frequency were varied symmetrically on the left and right wings. Under ideal bilateral symmetry, the roll- and yaw-moment contributions cancel, whereas pitch contributions may add; residual lateral and directional moments can arise from manufacturing, wing, or actuator asymmetries.} %Hence, the result suggests that amplitude and frequency modulation primarily regulate flight speed rather than orientation, while also potentially improving maneuverability by enabling larger accelerations during rapid maneuvers such as sharp turns.

\rev{Stroke-offset modulation provided clearer attitude-control authority. Increasing the symmetric offset $\delta_s=(\delta_L+\delta_R)/2$ (upward shift in neutral wing position as shown in Fig.~\ref{fig:Fig3_S3M}A) reduced the resultant force and changed the cycle-averaged pitch moment monotonically (Fig.~\ref{fig:Fig4_FT}F). This generated nose-down pitching torques, thereby confirming a functional pitch control channel establishing its control sign convention. Differential offset $\delta_d=(\delta_R-\delta_L)/2$ (antisymmetric shift in neutral wing position) produced a pronounced roll moment and a minimal total force and yaw moment response (Fig.~\ref{fig:Fig4_FT}G), while free-flight maneuvers produced repeatable heading changes. We therefore treat differential offset as an experimentally effective coupled roll–yaw directional channel, and the detailed aerodynamic sequence requires phase-resolved aerodynamic or flow measurements.} 

\begin{figure}[H]
	\centering
	\includegraphics[width=0.7\textwidth]{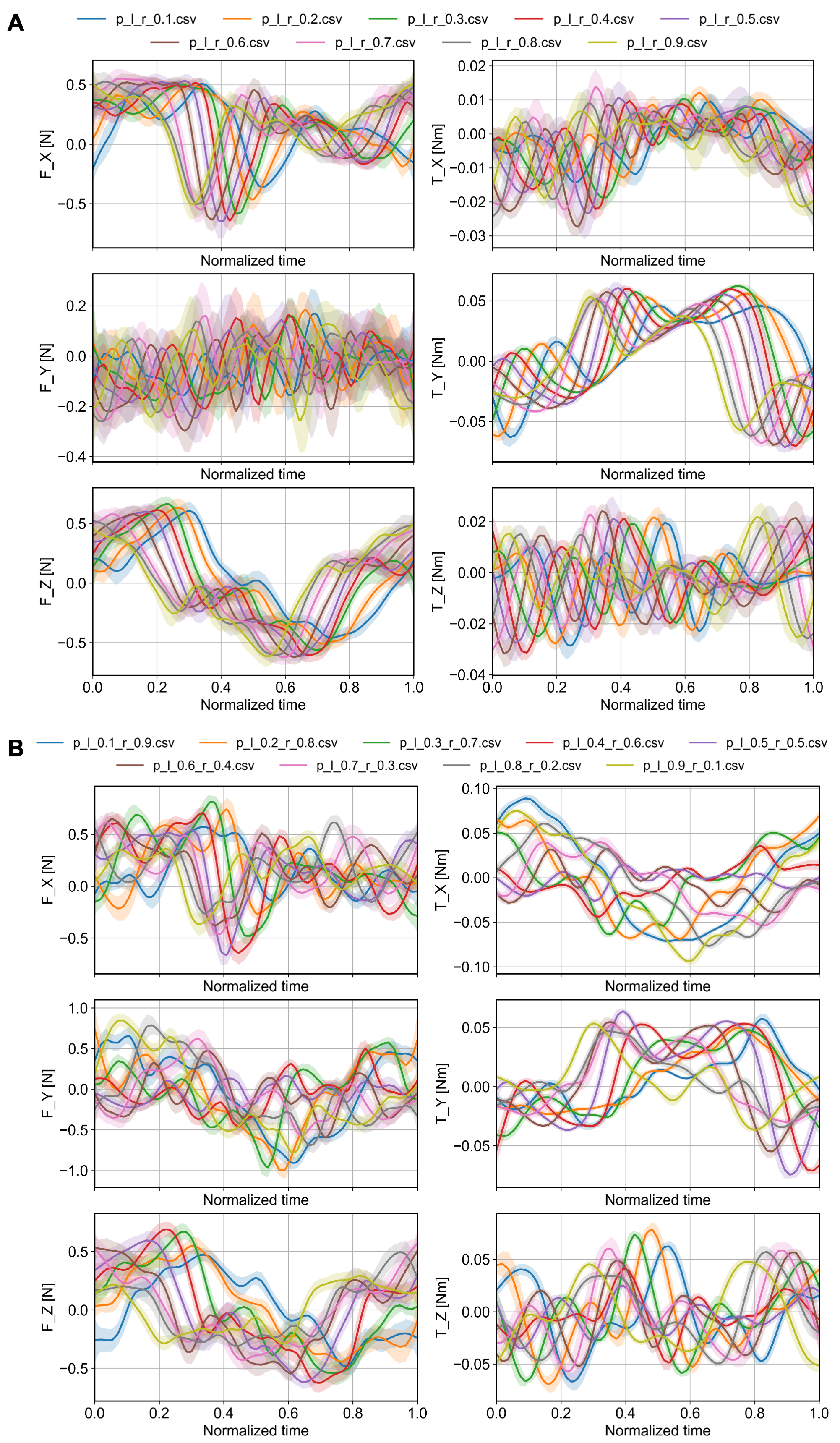} 
    \caption{\textbf{Forces and torques generated by stroke-timing modulation.} \rev{(\textbf{A}) Symmetric modulation (\texttt{p\_l\_r\_X}), for which both wings are assigned the same stroke-timing value $X$. The corresponding STAR parameters are $A_L=A_R=X-0.5$, and therefore the differential modulation is $\Delta A=A_R-A_L=0$. (\textbf{B}) Antisymmetric modulation (\texttt{p\_l\_X\_r\_Y}), for which the left and right wings are assigned stroke-timing values $X$ and $Y$, respectively. The corresponding STAR parameters are $A_L=X-0.5$ and $A_R=Y-0.5$, giving $\Delta A=A_R-A_L=Y-X$. The numerical values of $X$ and $Y$ are specified in the legend, and colors denote the different modulation settings.}}
	\label{fig:SM_StrokeVelocityVsForceTorque}
\end{figure}

\begin{figure}[H]
	\centering
	\includegraphics[width=0.95\textwidth]{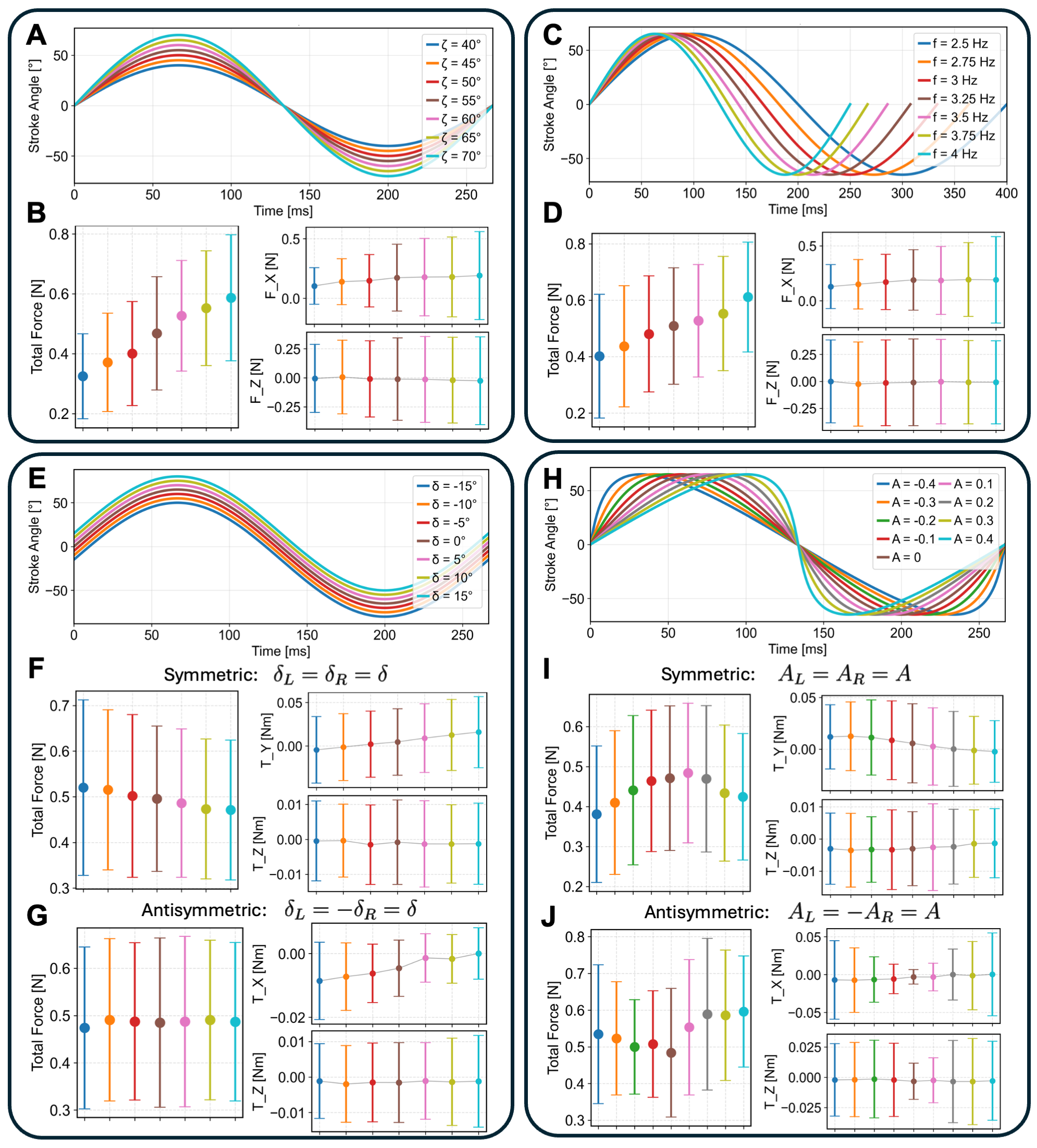} 
    \caption{\textbf{Force--torque mapping of flapping-wing modulation.} \rev{Each data point represents the temporal mean of the phase-averaged profile obtained from 40 phase-aligned wingbeats; error bars denote the standard deviation over wingbeat phase and therefore quantify intra-wingbeat load variation rather than uncertainty of the mean. The stroke-angle profiles are prescribed servo-command trajectories generated from the command laws without filtering or cycle averaging.} (\textbf{A}, \textbf{B}) Effect of flapping amplitude modulation: Increasing symmetric amplitude primarily increases the cycle-averaged body-$x$ force. (\textbf{C}, \textbf{D}) Effect of flapping frequency modulation: Increasing symmetric frequency increases body-$x$ force. (\textbf{E}--\textbf{G}) Effect of stroke-offset modulation: Symmetric stroke offset generates pitch moment, whereas differential offset produces a dominant roll-moment response with minimal yaw and force changes. (\textbf{H}--\textbf{J}) Effect of stroke-timing modulation: Symmetric STAR modulation alters pitch moment, whereas differential STAR modulation yields a dominant roll moment with a minor yaw response.}
	\label{fig:Fig4_FT} 
\end{figure}

STAR modulation produced a complementary set of wrench responses. Increasing the symmetric parameter \rev{$A_s=(A_L+A_R)/2$} (a faster downstroke relative to upstroke as shown in Fig.~\ref{fig:Fig3_S3M}B) monotonically shifted the cycle-averaged pitch moment, providing nose-up pitch-control authority (Fig.~\ref{fig:Fig4_FT}I). Differential STAR modulation \rev{$A_d=(A_R-A_L)/2$} (antisymmetric stroke timing between wings) yielded a pronounced roll moment with a minor yaw response, driving repeatable directional turns in free flight (Fig.~\ref{fig:Fig4_FT}J). \rev{Notably, STAR modulation produced larger intra-wingbeat variations in the phase-averaged force and moment profiles than stroke-offset modulation, indicating stronger phase-dependent redistribution of periodic loading (Supplementary Fig.~\ref{fig:SM_StrokeVelocityVsForceTorque}) rather than purely amplitude-driven changes (Supplementary Fig.~\ref{fig:SM_AngleOffsetVsForceTorque}).}

\rev{To connect measured force--torque responses to flight control allocation, individual wing inputs can be reconstructed relative to nominal operating values $(\delta_0, A_0)$ as $\delta_{L,R} = \delta_0+\delta_s \mp \delta_d$ and $A_{L,R} = A_0+A_s \mp A_d$. These empirical wrench trends guide control channel selection rather than serving as a complete aerodynamic model. In summary, amplitude and frequency are excluded from primary attitude control because their pitch moment variation was much smaller than that from stroke-offset and STAR modulation. Symmetric inputs ($\delta_s, A_s$) generate monotonic pitching moments, serving as longitudinal channels, whereas differential inputs ($\delta_d, A_d$) produce dominant lateral moments and clear free-flight heading changes, operating as directional channels. This empirical mapping constrains the feedback design to experimentally verified, monotonic control channels, with \ac{PID} gains tuned directly in free flight.}

\subsection*{Demonstration of undulatory free flight with attitude tracking}

Building on the experimentally determined control allocation, we developed an onboard control architecture comprising state estimation, attitude feedback control, and \ac{CPG} (Supplementary Fig.~\ref{fig:SM_ControllerArchitecture}). The Madgwick filter processes onboard \ac{IMU} measurements to provide attitude estimates at 100~Hz (Supplementary Appendix~\ref{app:pid}). \rev{For a controlled attitude variable $x\in\{\theta,\psi\}$, the feedback controller generates a virtual modulation command from the reference $x_{\mathrm{ref}}$ and estimate $\hat x$ using the controller gains $K_p, K_i, K_d$ as 
    \begin{equation} 
    \nu_x(t) = K_{p,x}e_x(t) + K_{i,x}\int_0^t e_x(\tau)\,\mathrm{d}\tau + K_{d,x}\dot e_x(t), \qquad e_x(t)=x_{\mathrm{ref}}(t)-\hat x(t). \label{eq:pid_virtual_modulation} 
    \end{equation} 
    The virtual command is then allocated to one experimentally selected modulation channel 
    \begin{equation} u_x(t) = \operatorname{sat} \left( g_x\nu_x(t); u_x^{\min}, u_x^{\max} \right), \label{eq:pid_control_allocation} \end{equation} where $u_\theta\in\{\delta_s,A_s\}, \, u_\psi\in\{\delta_d,A_d\}$. The sign $g_x\in\{1,-1\}$ is selected from the measured control-effectiveness relationship (Fig.~\ref{fig:Fig4_FT}, E to J), and the bounds $u_x^{\min}$ and $u_x^{\max}$ restrict the command to prevent mechanically inadmissible or untested wing motions. Only one of the two alternative channels (stroke-offset and STAR) is active in each reported flight experiment. The \ac{CPG} converts the resulting left- and right-wing commands ($\delta_L, \delta_R$) or ($A_L, A_R$) into independent periodic \ac{PWM} commands for the two servos. Thus, the force--torque mapping determines which \ac{CPG} parameter is modulated and in which direction, whereas the PID controller determines the time-varying modulation magnitude required to reduce the measured attitude error.} 
 
\begin{figure}[H]
	\centering
	\includegraphics[width=\textwidth]{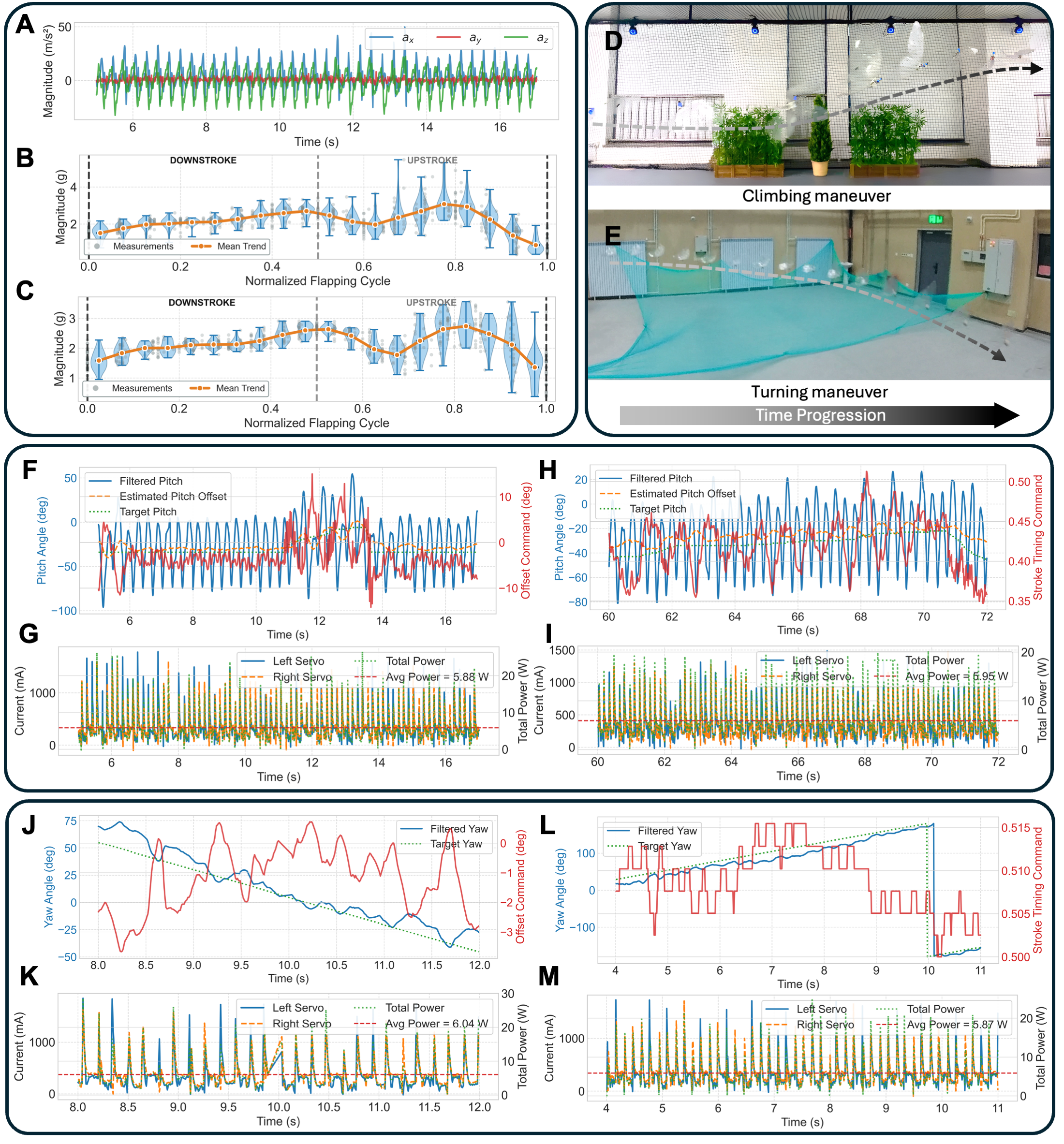}
	\caption{\textbf{Free-flight dynamics and closed-loop control.} 
(\textbf{A}) Tri-axial accelerometer measurements showing forward and vertical oscillatory accelerations during free flight. 
(\textbf{B}, \textbf{C}) Intra-cycle acceleration profiles under angle offset and stroke timing modulation, respectively. 
(\textbf{D}, \textbf{E}) Composite images of free-flight climbing and turning maneuvers. 
(\textbf{F}) Pitch angle tracking using angle offset modulation; RLS-extracted low-frequency mean enables stable feedback. 
(\textbf{H}) Pitch angle tracking using STAR stroke timing modulation with RLS filtering. 
(\textbf{J}, \textbf{L}) Yaw control for directional turning using angle offset and STAR modulation. 
(\textbf{G}, \textbf{I}, \textbf{K}, \textbf{M}) \rev{Corresponding measured mean power consumption during each maneuver.}
}
	\label{fig:Fig5_Control} 
\end{figure}

    As established previously, the AirPulse robot displays marked wingbeat-synchronous body dynamics at its approximately 3.25~Hz flapping frequency (Fig.~\ref{fig:Fig1_BiomimeticDesign}D and Fig.~\ref{fig:Fig5_Control}, F and H). \rev{During untethered flight, the measured pitch angle underwent wingbeat-synchronous excursions approaching $70^{\circ}$ (Fig.~\ref{fig:Fig5_Control}F). Onboard accelerometer recorded a mean forward-axis acceleration of $0.32\pm1.027~g$, a vertical-axis acceleration of $0.74\pm1.203~g$, and a total acceleration magnitude of $1.594\pm0.816~g$ with a measured peak of $5.475~g$ (Fig.~\ref{fig:Fig5_Control}A), where $g$ here denotes the standard gravitational acceleration ($9.81~\mathrm{m\,s^{-2}}$). These numbers reflect a strongly periodic body response driven by aggregate aerodynamic, inertial, structural, and actuator loads of flapping flight, rather than decomposing individual mechanisms or claiming kinematic equivalence with biological butterflies. Additionally, stroke-offset and STAR flights yielded qualitatively similar phase-resolved acceleration profiles (Fig.~\ref{fig:Fig5_Control}, B and C). Because the high-amplitude wingbeat directly introduces substantial periodic content into the instantaneous pitch measurement, feeding raw pitch angles into a low-bandwidth attitude loop would inject severe periodic content into the modulation commands.} Moreover, the wingbeat-synchronous body motion significantly weakens the timescale separation~\cite{taha2020vibrational} conventionally assumed in fast-flapping robots~\cite{keennon2012development,phan2017design,roshanbin2017colibri,kiani2019new}, thereby making a conventional stroke-averaged feedback approximation less appropriate. This necessitates explicit estimation of the slowly varying attitude component.

    An online \ac{RLS} estimator was therefore used to separate the dominant wingbeat harmonic from the slowly varying pitch component used for feedback (Methods; Fig.~\ref{fig:Fig5_Control}, F and H). Symmetric stroke-offset and STAR channels were evaluated for pitch tracking, and the corresponding differential channels were evaluated for directional tracking (Fig.~\ref{fig:Fig5_Control}, F, H, J, and L; Supplementary Movie~S4 and S5). Both pitch channels demonstrated reference-tracking behavior that validated the pitching moment signs identified in prior force--torque characterization, just as both differential channels successfully drove directional heading following. The measured mean power consumptions were 5.88 and 5.95~W for the two pitch trials, and 6.04 and 5.87~W for the two directional trials (Fig.~\ref{fig:Fig5_Control}, G, I, K, and M). These flight experiments validate the integration of onboard state estimation, empirically identified control channels, and rhythmic command generation for closed-loop untethered flight despite pronounced wingbeat-synchronous body undulations (Fig.~\ref{fig:Fig5_Control}, D and E).

    \rev{To complement the flight demonstrations, we identified local continuous-time \ac{OE} models for four representative trials, i.e., pitch and yaw tracking using stroke offset and STAR. For each trial, the model maps attitude reference to estimated attitude output and therefore represents the aggregate local closed-loop response, including the onboard controller, actuators, compliant structure, aggregate aerodynamics and inertia, state estimation, and delay (Methods). The pitch stroke-offset and STAR models reproduced the selected data with fits of $55.77\%$ and $65.58\%$ and RMSEs of $3.398^{\circ}$ and $1.974^{\circ}$, respectively. Their poles were $-30.00$, $-2.244$, and $-0.0768$ for stroke offset and $-1.78$ and $-0.2153\pm6.529\,\mathrm{i}$ for STAR. The yaw offset and STAR models had fits of $73.57\%$ and $94.25\%$ and RMSEs of $5.592^{\circ}$ and $3.114^{\circ}$, respectively, with poles at $-0.8227$ and $-0.01533\pm2.751\,\mathrm{i}$ for stroke offset and at $-0.7071$ and $-0.3423\pm1.723\,\mathrm{i}$ for STAR. Complete zeros, delays, and model orders are reported in Table~\ref{tab:closed_loop_identification}. All identified poles lie in the open left-half plane, so the fitted local models are asymptotically stable and reproduce the measured responses with moderate-to-high fit over the selected intervals. Notably, this analysis serves merely as a stability-consistency check for the tested closed-loop trajectories, not an open-loop plant identification, a global nonlinear model, or a stability proof for the time-periodic system.}
    
    \begin{table*}
    \centering 
    \rev{\caption{\textbf{Closed-loop OE identification of representative untethered flight trials.} These models describe the local reference-to-attitude mapping and only characterize local closed-loop tracking dynamics. \label{tab:closed_loop_identification} }} 
    \resizebox{\textwidth}{!}{% 
    \begin{tabular}{llcccccc} \toprule Reference & Modulation & Model order & Fit (\%) & RMSE ($^\circ$) & Zeros  & Poles   \\ \midrule 
    Pitch & Angle offset & 3 poles / 2 zeros & 55.77 & 3.398 & $-3.605,\;-0.0861$ & $-30.00,\;-2.244,\;-0.0768$  \\ 
    Pitch & STAR & 3 poles / 2 zeros & 65.58 & 1.974 & $-2.556,\;14.75$ & $-1.78,\;-0.2153\pm6.529\,\mathrm{i}$  \\ 
    Yaw & Angle offset & 3 poles / 2 zeros & 73.57 & 5.592 & $-1.442,\;-0.4998$ & $-0.8227,\;-0.01533\pm2.751\,\mathrm{i}$  \\ 
    Yaw & STAR & 3 poles / 1 zero & 94.25 & 3.114 & $-2.081$ & $-0.7071,\;-0.3423\pm1.723\,\mathrm{i}$ \\ \bottomrule 
    \end{tabular}% 
    } 
    \end{table*}

\section*{Methods}\label{sec11}

\subsection*{Fabrication of robot components}
\label{sec:method_robot}
The fuselage and wing-servo connectors were 3D printed in polyethylene terephthalate glycol (PETG) using fused-deposition modeling. PETG was selected over polylactic acid (PLA) and thermoplastic polyurethane (TPU) for its combination of rigidity and impact resistance, allowing the structure to endure cyclic inertial loads during sustained flapping.

Wings were fabricated from Mylar film reinforced with carbon fiber spars (a 1.0~mm main spar, 0.8~mm leading edge and hindwing contour, and 0.5~mm longitudinal veins). The carbon reinforcement provided a high stiffness-to-weight ratio and created a spatially nonuniform reinforcement architecture that permitted passive deformation of the more lightly supported membrane regions. Alternative membranes were evaluated: TPU exhibited excessive compliance and poor adhesion, while silicone-coated polyamide (PA66) offered durability but imposed excessive mass. All tested membranes were flight-capable; however, PET was ultimately chosen for its balance of light weight, stiffness, tear resistance, adhesive compatibility, and optical transparency for motion capture. Wing fabrication involved laser-printing a venation template, overlaying the PET film, and bonding carbon rods using UV-curable adhesive and cyanoacrylate. Left-right wing pairs were mass-balanced within 0.05~g ($\le1\%$ of a single wing mass) to minimize asymmetries that could induce unbalanced aerodynamic forces or fuselage vibrations from uneven stroke timing or deformation~\cite{de2018quad}.

Commercial micro servos (BlueArrow D30T MG HV, 4.6~g each) were selected for their high torque-to-weight ratio and metal geartrain, minimizing backlash under dynamic loading. Metal servo arms were used to prevent torsional deformation observed with plastic arms, which degraded stroke symmetry and amplitude. Although lighter custom actuators with planetary gearboxes could further reduce mass, off-the-shelf servos were preferred for their reliability, reproducibility, and ease of integration.

The onboard embedded system is based on an ESP32-S3-WROOM-1-N16R8 (dual-core 32-bit Xtensa LX7, 240~MHz, 16~MB Flash, 8~MB PSRAM), supporting parallel real-time wing actuation and wireless communication with a custom ground control station for data logging and monitoring. Power is supplied by 1S or 2S LiPo batteries through a multi-stage regulation system: a buck converter provides 3.3~V to the controller and sensors, a boost circuit delivers 8.2~V to the servos for consistent torque under high load, and an LDO outputs 5~V to the ExpressLRS module. The ELRS module serves only as a safety override for manual control. Servo current and power consumption are monitored via INA219 sensors. The sensor suite includes an ICM-42688-P IMU (±16~g, ±2000~°/s, 32~kHz), BMM350 magnetometer (±2000~$\mu$T, 400~Hz), and BMP390L barometer (±3~Pa, 200~Hz) for real-time state estimation.

\subsection*{Forewing contour parameterization}
\label{sec:method_forewing_param}
Sampled wing contour (data points $\{\mathbf{D}_j\}_{j=1}^N$) were represented by a periodic cubic B-spline. The spline curve is defined as
\begin{equation}
    \mathbf{C}(u)=\sum_{i=0}^{n-1} N_{i,3}(u)\,\mathbf{P}_i,\qquad u\in[0,1], 
    \label{eq:spline_curve}
\end{equation}
where $\{\mathbf{P}_i\}_{i=0}^{n-1}$ are the spline control points determined by smoothing least-squares fitting, and $N_{i,3}(u)$ are the cubic B-spline basis functions defined by the Cox-deBoor recursion~\cite{de1978practical} as
\begin{align}
    N_{i,0}(u)&=
\begin{cases}
1,& u_i\le u < u_{i+1},\\
0,&\text{otherwise}, 
\end{cases} \\
N_{i,p}(u)&=\frac{u-u_i}{u_{i+p}-u_i}N_{i,p-1}(u)+\frac{u_{i+p+1}-u}{u_{i+p+1}-u_{i+1}}N_{i+1,p-1}(u),
\label{eq:cox_de_boor}
\end{align}
with the convention that any fraction with a zero denominator is taken as zero. A periodic knot vector $\mathbf{U}$ was used to ensure closure and $C^2$ continuity.

Raw contours were cyclically reordered to start at the wing root and closed by appending the start point (Supplementary Fig.~\ref{fig:SM_WingContourParameterization}A). Parameter values were assigned by chord length as
\begin{equation}
    t_k=\frac{\sum_{m=1}^{k-1}\|\mathbf{D}_{m+1}-\mathbf{D}_m\|}{\sum_{m=1}^{N}\|\mathbf{D}_{m+1}-\mathbf{D}_m\|},\quad t_1=0,\; t_{N+1}=1,
    \label{eq:chord_length}
\end{equation}
and a periodic smoothing cubic spline was fitted to $(t_k,\mathbf{D}_k)$ by least-squares with smoothing factor $s=\alpha_\mathrm{s} N$ ($\alpha_\mathrm{s}=0.5$ herein). The fitted spline was resampled at $M=1000$ uniformly spaced $u\in[0,1]$ to yield a dense curve $\widetilde{\mathbf{C}}$. Fit quality was quantified by nearest-neighbor errors as
\begin{equation}
    e_j=\min_{q\in\widetilde{\mathbf{C}}}\|\mathbf{D}_j-q\|,
\qquad
\mathrm{RMS}=\sqrt{\frac{1}{N}\sum_{j=1}^N e_j^2}.
\label{eq:error_metric}
\end{equation}

The parameterization results from the above analysis were reported in Table~\ref{tab:parameterization_results} and the control points $\mathbf{P}_i = (x_i, y_i)$ for the forewing spline representation were summarized in Table~\ref{tab:control_points}. The fitting procedure achieved an RMS error of 0.672 pixels (Supplementary Fig.~\ref{fig:SM_WingContourParameterization}B), corresponding to a relative error of 0.13\% of the wing span. The error distribution (Supplementary Fig.~\ref{fig:SM_WingContourParameterization}C) shows that 95\% of contour points were fitted with errors less than 1.5 pixels, demonstrating the high accuracy of the B-spline representation while maintaining $C^2$ continuity. This parameterization procedure produced smooth, reproducible wing contours suitable for CAD and aeroelastic analysis, with the compact control point representation enabling computational modeling while \rev{preserving geometric fidelity to the reference wing contour}.

\subsection*{Biological wing venation analysis}
\label{sec:method_wing_venation}

Forewing images were collected primarily from published literature with documented clear venation patterns~\cite{patil2017insight}, supplemented by photographs of museum-grade specimens available from open-access sources (Wikipedia). Specimens were selected based on strict visibility criteria requiring a fully unobstructed forewing (dorsal or ventral) view with distinct Dc and Cu1 veins, which define key geometric axes for comparative analysis. The dense and pigmented scales on butterfly wings often obscure vein boundaries, substantially constraining the number of usable samples. Consequently, the final dataset included 24 different specimens representing three families, namely Nymphalidae ($n=8$), Pieridae ($n=8$), and Papilionidae ($n=8$). For each specimen, the forewing image was first converted to grayscale and manually annotated with yellow solid lines indicating the Dc and Cu1 veins (Supplementary Fig.~\ref{fig:SM_WingStudies}). The annotated regions were then segmented in HSV color space to extract the obtuse inter-vein Dc-Cu1 angle. Future morphological studies could expand statistical results by employing chemical bleaching or depigmentation techniques to enhance venation contrast and enable automated image segmentation.

\subsection*{\rev{Characterization of compliant wings}}
\label{sec:method_compliant_wing_characterization}

The characterization of compliant wings was performed using a custom flapping apparatus (Fig.~\ref{fig:Fig2_Characteristics}D) equipped with a six-axis load cell and a retroreflective-marker-based motion-capture system (LUSTER FZMotion). Lightweight retroreflective markers were used to minimize their influence on wing mass distribution and deformation. \rev{Test apparatus was rigidly clamped to suppress mechanical vibration.} Each wing variant was subjected to the same commanded flapping motion, while synchronized force and kinematic data were recorded at $222~\mathrm{Hz}$. Active flapping intervals were identified from the servo \ac{PWM} commands, and individual wingbeats were segmented using the measured kinematics. For each configuration, 40 consecutive cycles during steady periodic operation were retained. Force signals were processed using a fourth-order Butterworth low-pass filter with a $50~\mathrm{Hz}$ cutoff. \rev{Each wingbeat was then mapped onto the normalized phase $\tau=t/T\in[0,1]$ and phase-aligned using the measured kinematics. The phase-averaged waveform was obtained by averaging the aligned signals pointwise across the 40 cycles at each normalized phase.
% , $\bar{\mathbf{f}}(\tau) = \frac{1}{N}\sum_{k=1}^{N}\mathbf{f}_k(\tau), \, N=40$. 
This procedure preserves repeatable phase-dependent loading while attenuating cycle-incoherent measurement noise and cycle-to-cycle variability.}

\rev{To estimate the aerodynamic contribution, the approximate aerodynamic force $\mathbf{f}_{\mathrm{aero},j}(\tau)$ for the $j$-th configuration was calculated by subtracting the baseline inertial measurement from the full-wing measurement, \begin{equation} \mathbf{f}_{\mathrm{aero},j}(\tau) \approx \mathbf{f}_{\mathrm{full},j}(\tau) - \mathbf{f}_{\mathrm{inertial},j}(\tau), \label{eq:aero_force} \end{equation} where $\mathbf{f}_{\mathrm{full},j}$ and $\mathbf{f}_{\mathrm{inertial},j}$ denote the phase-averaged load-cell measurements obtained with the complete wing and the membrane-removed structural baseline, respectively. Equation~\eqref{eq:aero_force} therefore provides an estimate of aerodynamic loading rather than an exact decomposition, owing to fluid--structure coupling, small differences in mass distribution, and residual aerodynamic loading on the exposed reinforcement structure.} Processed forces were further represented in polar coordinates, with angular position denoting normalized wingbeat phase and radial position denoting force magnitude with an offset to ensure positivity (Fig.~\ref{fig:Fig2_Characteristics}, F and G). Time-integrated forces were computed using trapezoidal numerical integration for both signed and absolute components, providing measures of net impulse and cumulative force production, respectively.

\rev{To enable area- and velocity-normalized comparisons across wing configurations, the estimated aerodynamic forces were further expressed as nondimensional force coefficients. For wing configuration $j$, the radius of the second moment of
wing area was defined as
\begin{equation}
r_{2,j}
=
\left(
\frac{1}{S_j}
\int_{S_j} r^2 \, dS
\right)^{1/2}
=
R_j \hat{r}_{2,j},
\label{eq:second_moment_radius}
\end{equation}
where $S_j$ is the planform area of wing configuration $j$, $r$ is the radial distance of an elemental area $dS$ from the flapping axis, $R_j$ is the maximum wing radius, and $\hat{r}_{2,j}$ is the nondimensional radius of the second
moment of wing area. For full-wing and forewing-only configuration, $r_2$ is computed as $0.128~\mathrm{m}$ and $0.152~\mathrm{m}$, respectively. Then, assuming sinusoidal flapping, \begin{equation}
\phi_j(t)
=
\phi_{0,j}
+
\frac{\Phi_j}{2}
\sin\left(2\pi f_j t\right),
\label{eq:sinusoidal_flapping_motion}
\end{equation}
where $\phi_j(t)$ is the instantaneous flapping angle, $\phi_{0,j}$ is its mean value, $\Phi_j$ is the peak-to-peak flapping amplitude in radians, and $f_j$ is the flapping frequency. The corresponding characteristic velocity was defined
using the radius of the second moment of wing area as
\begin{equation}
U_{\mathrm{ref},j}
=
2 \Phi_j f_j R_j \hat{r}_{2,j}
=
2 \Phi_j f_j r_{2,j}.
\label{eq:reference_velocity}
\end{equation}
For force component $k \in \{x,y,z\}$, the instantaneous nondimensional force
coefficient was then calculated as
\begin{equation}
C_{F_k,j}(\tau)
=
\frac{
\widehat{f}_{\mathrm{aero},k,j}(\tau)
}{
\frac{1}{2}
\rho
U_{\mathrm{ref},j}^{2}
S_j
},
\label{eq:instantaneous_force_coefficient}
\end{equation}
where $\widehat{f}_{\mathrm{aero},k,j}(\tau)$ denotes the estimated aerodynamic force obtained by~\eqref{eq:aero_force}, $\rho$ is the air density, and $\tau=t/T_j \in [0,1]$ is the normalized wingbeat phase, with $T_j=1/f_j$. The signed cycle-averaged force coefficient was subsequently computed as
\begin{equation}
\overline{C}_{F_k,j}
=
\int_{0}^{1}
C_{F_k,j}(\tau)
\, d\tau.
\label{eq:cycle_averaged_force_coefficient}
\end{equation}
The dimensional forces were retained in parallel to quantify the absolute force-generating capacity available to the robot, whereas $\overline{C}_{F_k,j}$ provides an area- and velocity-normalized comparison among wing configurations. To compute~\eqref{eq:instantaneous_force_coefficient}, the reference velocities for the full-wing and forewing-only configurations were determined as \(U_\mathrm{ref, full} = 1.308~\mathrm{m\,s^{-1}}\) and \(U_\mathrm{ref, fore} = 1.550~\mathrm{m\,s^{-1}}\), respectively, based on \(\Phi = 1.5708~\mathrm{rad}\), \(f = 3.25~\mathrm{Hz}\), and the previously calculated \(r_{2}\). The remaining parameters were set to an air density of \(\rho = 1.225~\mathrm{kg\,m^{-3}}\), with corresponding wing areas of \(S_\mathrm{full} = 0.0549~\mathrm{m^2}\) for the full-wing and \(S_\mathrm{fore} = 0.0344~\mathrm{m^2}\) for the forewing-only configuration.}

Wing deformation was quantified by projecting three-dimensional marker trajectories onto a plane orthogonal to the wing rotation axis, thus decoupling bending deformation from the primary flapping motion. Bending angles were defined as the angular deviation between the proximal and distal wing segments, with maxima identified separately for the downstroke and upstroke (Fig.~\ref{fig:Fig2_Characteristics}, I and J). Hindwing kinematics were characterized by computing elevation angles relative to the wing hinge throughout each flapping cycle. Forewing kinematics were derived from markers placed along the anterior (leading edge) and posterior (trailing edge) margins of each wing section, enabling direct quantification of instantaneous wing orientation relative to the airflow. Discrete marker data were interpolated to generate continuous angle-time curves, enabling phase-resolved comparison of fore- and hind-wing kinematics across wing variants.

\subsection*{Analytical properties of STAR}
\label{sec:method_STAR_properties}

\paragraph{(i) Cycle-averaged frequency invariance.} For constant $A$, the duration of one complete wingbeat is 
\begin{equation} T = \int_0^{2\pi}\frac{d\omega}{\dot{\omega}} = \frac{1}{\pi f}\int_0^{2\pi}p(\omega)\,d\omega = \frac{1}{f}, 
\end{equation} 
since $\int_0^{2\pi}\cos\omega\,d\omega=0$. Thus, STAR redistributes time within each wingbeat without altering the cycle-averaged flapping frequency.

\paragraph{(ii) Linear stroke asymmetry.} Choosing the phase origin such that the downstroke corresponds to \(\omega \in [-\pi/2, \pi/2]\) and the upstroke to \(\omega \in [\pi/2, 3\pi/2]\), the half-stroke durations are
\begin{align} T_{\mathrm{down}} &=\frac{1}{\pi f}\int_{-\pi/2}^{\pi/2}p(\omega)\,d\omega =\frac{1}{2f}\rev{-}\frac{2A}{\pi f},\\ T_{\mathrm{up}} &=\frac{1}{\pi f}\int_{\pi/2}^{3\pi/2}p(\omega)\,d\omega =\frac{1}{2f}\rev{+}\frac{2A}{\pi f}. \end{align} 
\rev{Hence, 
\begin{equation} \frac{T_{\mathrm{up}}-T_{\mathrm{down}}} {T_{\mathrm{up}}+T_{\mathrm{down}}} =\frac{4A}{\pi}, \end{equation} 
showing that stroke-timing asymmetry varies linearly and monotonically with $A$: positive $A$ shortens the downstroke and prolongs the upstroke, corresponding to a faster downstroke relative to the upstroke.}

\paragraph{(iii) Smoothness and boundedness.} Since $p(\omega)=0.5-A\cos\omega$ is smooth and strictly positive for $|A|<0.5$, its reciprocal 
\begin{equation} r(\omega)=\frac{1}{p(\omega)}, \qquad \dot{\omega}=\pi f\,r(\omega), \end{equation} 
is continuous and bounded, with 
\begin{equation} \frac{\pi f}{0.5+|A|} \leq \dot{\omega} \leq \frac{\pi f}{0.5-|A|}. 
\end{equation} 
Because $y=\zeta\sin\omega$ and $\dot{\omega}$ are continuous, both the commanded stroke angle $y$ and its first derivative $\dot{y}=\zeta\cos\omega\,\dot{\omega}$ remain continuous and bounded.

\subsection*{\rev{Discrete-time} implementation of STAR}
\label{sec:method_STAR_implementation}

To suppress abrupt changes in phase speed when the modulation parameter is updated, the reciprocal phase-speed variable is low-pass filtered using a first-order \ac{IIR} scheme. For servo-loop index $n$ and update rate $f_{\mathrm{servo}}$, \begin{align} r_{\mathrm{target}}[n] &=\frac{1}{0.5-A[n]\cos\omega[n]},\\ r_{\mathrm{smooth}}[n] &=\alpha_{\mathrm{s}}r_{\mathrm{target}}[n] +(1-\alpha_{\mathrm{s}})r_{\mathrm{smooth}}[n-1],\\ \Delta\omega[n] &=\frac{\pi f}{f_{\mathrm{servo}}}\, r_{\mathrm{smooth}}[n],\\ \omega[n+1] &=\omega[n]+\Delta\omega[n], \end{align} where $\alpha_{\mathrm{s}}\in(0,1)$ trades responsiveness against kinematic smoothness. For a desired first-order cutoff frequency $f_c$, \begin{equation} \alpha_{\mathrm{s}} =1-\exp\!\left(-\frac{2\pi f_c}{f_{\mathrm{servo}}}\right) \approx \frac{2\pi f_c}{f_{\mathrm{servo}}}, \end{equation} where the approximation holds for $f_c\ll f_{\mathrm{servo}}$. The cutoff is selected below the flapping frequency and further tuned experimentally to balance attitude-response agility against wingstroke smoothness.

\subsection*{Online pitch estimation using recursive least squares}
\label{sec:method_RLS}

To estimate the slowly varying pitch component in the presence of wingbeat-synchronous motion, the measured pitch signal can be modeled as
\begin{equation}
    y(t) = a \sin(\omega t) + b \cos(\omega t) + c + \epsilon(t),
\end{equation}
where $a$ and $b$ describe the flapping-induced sinusoidal component at frequency $\omega$, $c$ is the low-frequency mean, and $\epsilon(t)$ represents measurement noise.

In discrete time with sampling index $n$ and step $\Delta t$, the regression vector is
\begin{equation}
    \boldsymbol{\varphi}_n =
    \begin{bmatrix}
        \sin(\omega t_n)\\
        \cos(\omega t_n)\\
        1
    \end{bmatrix}, \quad 
    y_n = \boldsymbol{\varphi}_n^\top \boldsymbol{\theta} + \epsilon_n, \quad 
    \boldsymbol{\theta} = [a, b, c]^\top,
\end{equation}
and the RLS updates follow
\begin{align}
    \boldsymbol{K}_n &= \frac{\boldsymbol{P}_{n-1} \boldsymbol{\varphi}_n}{\lambda + \boldsymbol{\varphi}_n^\top \boldsymbol{P}_{n-1} \boldsymbol{\varphi}_n} \quad (\text{gain vector}),\\
    e_n &= y_n - \boldsymbol{\varphi}_n^\top \boldsymbol{\theta}_{n-1}\quad (\text{prediction error}),\\
    \boldsymbol{\theta}_n &= \boldsymbol{\theta}_{n-1} + \boldsymbol{K}_n e_n\quad (\text{parameter update}),\\
    \boldsymbol{P}_n &= \frac{1}{\lambda}\left[\boldsymbol{P}_{n-1} - \boldsymbol{K}_n \boldsymbol{\varphi}_n^\top \boldsymbol{P}_{n-1}\right]\quad (\text{covariance update}),
\end{align}
where $\lambda \in (0,1]$ is the forgetting factor controlling adaptation speed. The low-frequency pitch signal for inner-loop control is obtained directly from the updated estimate $c_n$.

Furthermore, the regressor can be adapted using an instantaneous flapping phase to account for variations in flapping frequency, 
\begin{equation}
    \varphi(t_n) = \int_0^{t_n} \omega_{\text{inst}}(\tau)\,d\tau,
\end{equation}
yielding
\begin{equation}
    \boldsymbol{\varphi}_n^{\text{adaptive}} =
    \begin{bmatrix}
        \sin(\varphi(t_n))\\
        \cos(\varphi(t_n))\\
        1
    \end{bmatrix}.
\end{equation}
This frequency-aware regressor ensures robust tracking of the pitch mean under stroke timing modulation (with the instantaneous phase calculated from \eqref{eq:S3M}).

\rev{\subsection*{Closed-loop system identification and stability analysis}
\label{sec:closed_loop_identification}}

\rev{Closed-loop dynamics were identified from representative untethered flight experiments, including pitch tracking using stroke-angle offset modulation, pitch tracking using STAR modulation, yaw tracking using stroke-angle offset modulation, and yaw tracking using STAR modulation. Let $x\in\{\theta,\psi\}$ denote the controlled attitude variable, where $\theta$ is pitch angle and $\psi$ is yaw angle. For each experiment, we extracted the attitude reference $r_x(t)$, the estimated attitude $y_x(t)$ used by the onboard controller, and the corresponding modulation command $u_x(t)$. For an input $v_x(t)\in\{u_x(t),r_x(t)\}$, a continuous-time output-error model was defined as
\begin{equation}
M_{x,v}(s;\boldsymbol{\vartheta})
=
\frac{B(s;\boldsymbol{\vartheta})}{A(s;\boldsymbol{\vartheta})}
=
\frac{b_0s^m+b_1s^{m-1}+\cdots+b_m}
{s^n+a_1s^{n-1}+\cdots+a_n},
\qquad m<n,
\label{eq:oe_transfer_function}
\end{equation}
where $n\in\{2,3\}$ and $m\in\{1,2\}$ are the candidate numbers of poles and zeros, respectively. The model was simulated in the time domain with zero-order-held input samples, an integer input delay $d\geq0$, and an estimated initial state $\mathbf{x}_0$. The predicted output was written as
\begin{equation}
\hat{y}_x(t_k)
=
y_0
+
\mathcal{S}
\left[
M_{x,v}(s;\boldsymbol{\vartheta}),
\, v_x(t_k)-v_0,
\, d,
\, \mathbf{x}_0
\right],
\label{eq:raw_offset_oe_prediction}
\end{equation}
where $v_0$ and $y_0$ are the fitted input and output operating-point offsets, and $\mathcal{S}[\cdot]$ denotes continuous-time simulation of the transfer function, including the delayed forced response and free response from $\mathbf{x}_0$. Thus, the reported transfer function describes the local perturbation map from $v_x-v_0$ to $y_x-y_0$, while the offsets are inferred from the same raw flight segment rather than imposed by sample-mean removal.}

\rev{For each candidate model, the parameters, offsets, initial state, and delay were estimated by solving
\begin{equation}
\left\{
\boldsymbol{\vartheta}^{*},
v_0^{*},
y_0^{*},
\mathbf{x}_0^{*},
d^{*}
\right\}
=
\operatorname*{arg\,min}_{\boldsymbol{\vartheta},\,v_0,\,y_0,\,\mathbf{x}_0,\,d}
\sum_{k=1}^{N}
\rho
\left[
y_x(t_k)-\hat{y}_x(t_k)
\right],
\label{eq:oe_identification_objective}
\end{equation}
where $\rho(e)=2(\sqrt{1+e^2}-1)$ is a soft-$L_1$ loss. Candidate orders and delays were pre-screened using discrete-time autoregressive models with exogenous input, after which the continuous-time output-error candidates were ranked using the Bayesian information criterion. The final reported transfer functions were always obtained from the raw-offset continuous-time output-error fits.}

\rev{Two input--output mappings were considered. Setting $v_x=u_x$ gives the local command-response model
\begin{equation}
G_{u,x}(s)=\frac{\Delta Y_x(s)}{\Delta U_x(s)},
\qquad
\Delta u_x=u_x-u_0,\quad
\Delta y_x=y_x-y_0.
\end{equation}
Because $u_x(t)$ is generated by feedback from the tracking error, this model is a closed-loop command-response approximation and cannot, by itself, be interpreted as the unique open-loop plant. Setting $v_x=r_x$ gives the closed-loop tracking model
\begin{equation}
T_x(s)=\frac{\Delta Y_x(s)}{\Delta R_x(s)},
\qquad
\Delta r_x=r_x-r_0,
\end{equation}
which was used for stability-consistency analysis. Model accuracy was quantified using
\begin{equation}
\operatorname{RMSE}
=
\left[
\frac{1}{N}
\sum_{k=1}^{N}
\left(y_x(t_k)-\hat{y}_x(t_k)\right)^2
\right]^{1/2},
\end{equation}
and
\begin{equation}
\operatorname{Fit}_{\%}
=
100
\left(
1-
\frac{\left\|\mathbf{y}_x-\hat{\mathbf{y}}_x\right\|_2}
{\left\|\mathbf{y}_x-\overline{y}_x\mathbf{1}_N\right\|_2}
\right),
\qquad
\overline{y}_x=\frac{1}{N}\sum_{k=1}^{N}y_x(t_k).
\end{equation}
Writing $T_x(s)=N_x(s)/D_x(s)$, the identified closed-loop poles are the roots satisfying $D_x(p_i)=0$. The local linear tracking model is asymptotically stable when $\operatorname{Re}(p_i)<0$ for all $i$. This analysis provides a model-based consistency check for local closed-loop stability under the tested operating conditions, rather than a formal stability proof of the global nonlinear time-periodic flapping-wing dynamics.}

\section*{Discussion}\label{sec12}

\rev{The primary contribution of this study is the integrated development and experimental validation of onboard attitude control for a 26~g two-winged tailless flapping-wing robot with pronounced wingbeat-synchronous dynamics. The platform, modulation-to-wrench characterization, STAR formulation, and onboard controller are mutually dependent elements of this system-level result. The compliant venation-informed wings and swallowtail-derived planform provide morphological and structural design cues, whereas the commanded kinematics, force--torque mapping, STAR modulation law, state estimation, and feedback control are engineering developments. Prior platforms, notably Festo's eMotionButterflies~\cite{festo2015emotionbutterfly}, demonstrated untethered flight using external tracking and offboard control. The AirPulse robot differs by executing state estimation, rhythmic command generation, and closed-loop attitude control onboard. This distinction reflects control architecture rather than overall flight performance. Nevertheless, among butterfly-inspired platforms capable of sustained untethered flight (Table~\ref{tab:butterfly_comparison}), the AirPulse robot achieves the lowest reported flight-ready mass.}

\begin{figure}[H]
	\centering
	\includegraphics[width=\textwidth]{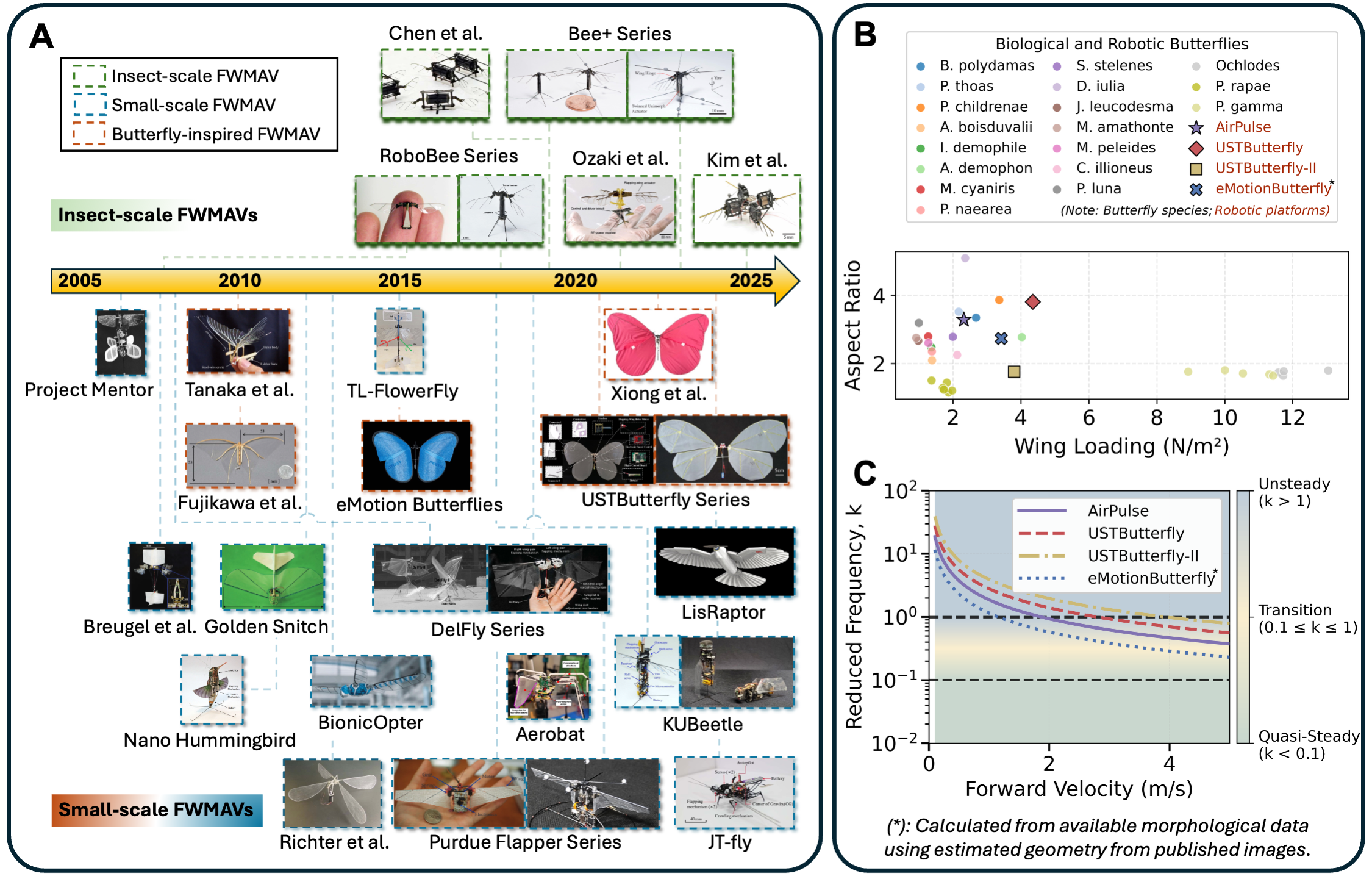} 
    \caption{\textbf{Historical context, \rev{morphological comparison}, and aerodynamic operating regime of the AirPulse robot.} (\textbf{A}) Timeline of selected \ac{FWMAV} development from 2005 to 2025, highlighting insect-scale (green), small-scale (blue), and butterfly-inspired (orange) platforms. (\textbf{B}) Comparison of aerodynamic aspect ratio, $\textrm{AR}=b^2/(2S)$, and wing loading, $\textrm{WL}=mg/(2S)$, across selected butterflies and butterfly-inspired robots (Table~\ref{tab:butterfly_comparison}). The AirPulse robot occupies a region similar to that of \textit{Papilio thoas} and has one of the lowest wing loadings among the robotic platforms considered. \rev{Lower wing loading reduces the weight supported per unit wing area and facilitates low-speed flight.} (\textbf{C}) \rev{Reduced-frequency estimate $k=\omega\bar c/(2U_\infty)$ for the AirPulse robot. Values above unity indicate the platform in a strongly unsteady operating regime in which transient aerodynamic effects are expected to be important.} }
	\label{fig:Fig7_Discussion} % g
\end{figure}

\rev{From a structural and scaling perspective, the AirPulse robot occupies a region of the AR--WL (aspect-ratio--wing-loading) plane similar to that of the swallowtail butterfly \textit{Papilio thoas} (Fig.~\ref{fig:Fig7_Discussion}B). Here, the aerodynamic aspect ratio and wing loading are defined as $\textrm{AR}=b^2/(2S), \, \textrm{WL}=mg/(2S)$, where $b$ is the total wingspan, $S$ is the planform area of one lateral wing assembly, $m$ is the flight-ready mass, and $g$ is gravitational acceleration. Wing loading has units of pressure and represents the robot weight supported per unit total wing area. The AirPulse robot has one of the lowest wing loadings among the butterfly-inspired robots considered in Table~\ref{tab:butterfly_comparison} ~\cite{huang2023ustbutterfly,huang2024aerodynamic}. For comparable air density and cycle-averaged lift coefficient, a lower wing loading reduces the characteristic aerodynamic velocity, thereby facilitating low-speed flight~\cite{nan2018can,kang2018experimental}. However, wing loading alone guarantees neither energetic efficiency nor maneuverability since increased wing area may introduce penalties in profile drag and structural mass while the total system performance relies heavily on flapping kinematics and actuator bandwidth. Consequently, we use the AR--WL diagram strictly to contextualize the morphology and loading scale of our designed robot. The characteristic operating regime of the AirPulse robot is governed by low-Reynolds-number and highly unsteady aerodynamic effects (Fig.~\ref{fig:Fig7_Discussion}C). The total effective velocity over the wing is calculated as $V_{\mathrm{total}} = \sqrt{U_{\mathrm{ref}}^2 + U_{\infty}^2} \approx 1.646~\mathrm{m\,s^{-1}}$, which combines the characteristic velocity due to flapping motion ($U_{\mathrm{ref}} = 1.308~\mathrm{m\,s^{-1}}$ as derived earlier) with the typical forward flight speed ($U_{\infty} = 1~\mathrm{m\,s^{-1}}$). This yields an estimated Reynolds number of $Re = V_{\mathrm{total}}\bar{c}/\nu \approx 20{,}637$, given a mean aerodynamic chord $\bar{c} = 183~\mathrm{mm}$ and an air kinematic viscosity $\nu = 1.46 \times 10^{-5}~\mathrm{m^2\,s^{-1}}$. 
% This regime is comparable to other bio-inspired flapping-wing platforms in the literature (Fig.~\ref{fig:Fig7_Discussion}A), for instance, DelFly Nimble~\cite{karasek2018tailless} and Bat Bot~\cite{ramezani2017biomimetic} ($Re \approx 52{,}300$ with $U_\infty=5.6~\mathrm{m\,s^{-1}}$ and $\bar{c}=0.14~\mathrm{m}$). 
Concurrently, the non-dimensional reduced frequency, defined as $k = \omega\bar{c}/(2U_{\infty}) \approx 1.87$ (where $\omega = 2\pi f$ is the angular flapping frequency at $f = 3.25~\mathrm{Hz}$), places the platform in a strongly unsteady operating regime in which transient aerodynamic effects are expected to be important.}

\rev{Despite these advances, several limitations and open challenges remain. First, while empirical \ac{PID} tuning effectively achieves pitch and yaw tracking, it lacks guaranteed robustness under dynamically varying aerodynamic loads or exogenous disturbances. The strong fluid--structure interaction intrinsic to flexible wings introduces complex nonlinearities unmodeled by classical control frameworks. Developing higher-fidelity, time-periodic dynamic models remains essential for designing adaptive or neural control architectures~\cite{gu2020uav}, as well as for rigorous stability analysis via high-order averaging~\cite{taha2020vibrational} or Floquet theory~\cite{saetti2022}. Second, while a venation-informed reinforcement layout is adopted in our wing design, the specific influence of interspecific vein topology and angles on aerodynamic performance remains unresolved. Existing biological datasets are insufficient to statistically discern whether natural venation variations reflect functional evolutionary adaptations or morphological drift. Addressing this fundamental question requires an interdisciplinary approach integrating biological morphology, computational fluid dynamics, and robotic experimentation. Moreover, systematic characterization of venation-informed, low-aspect-ratio wings remains an important direction for future work. Looking ahead, integrating neuromorphic architectures to couple event-driven vision~\cite{paredes2024fully} and olfactory sensing~\cite{deveza1994odor} will be crucial for autonomous navigation in cluttered or GPS-denied environments, advancing insect-scale sensorimotor integration.}

% \section{Conclusion}\label{sec13}

% Conclusions may be used to restate your hypothesis or research question, restate your major findings, explain the relevance and the added value of your work, highlight any limitations of your study, describe future directions for research and recommendations. 

% In some disciplines use of Discussion or 'Conclusion' is interchangeable. It is not mandatory to use both. Please refer to Journal-level guidance for any specific requirements. 

\backmatter

\bmhead{Supplementary information}
Movie S1: Overview of the AirPulse robot and demonstration of \rev{onboard closed-loop controlled} flight; Movie S2: \rev{Forewing--hindwing phase lag and passive feathering in the compliant wing design}; Movie S3: \rev{Qualitative comparison of wingbeat-synchronous body
motion in a butterfly and the AirPulse robot}; Movie S4: Demonstration of a climbing maneuver; Movie S5: Demonstration of a turning maneuver.

\bmhead{Acknowledgements}
We are grateful to Prof. G.C.H.E. de Croon, Prof. Maurizio Porfiri, and Prof. Lu Fang for their valuable comments and feedback on the early version of this manuscript. W.G. acknowledges support from the China Postdoctoral Science Foundation under Grant Number 2025M781650. During the preparation of this manuscript, the authors used GPT-5.5 to refine the language and improve readability. The authors reviewed and edited the output and take full responsibility for the content of the final manuscript.

\bmhead{Author contributions}
W.G. led the project, conceived the ideas, and designed the robots and test apparatus.
C.F. and X.J. designed the electronics.
L.L. and X.S. designed the mechanical structure.
W.G., C.F., C.Y., and Y.D. devised the control algorithm.
C.F. and Y.D. developed the ground control station and visualization software.
W.G., C.F., L.L., C.Y., X.J., Y.D., and X.S. performed the experiments.
W.G., C.F., and L.L. conducted data analysis.
W.G. and L.L. prepared the figures and movies.
W.G. wrote and revised the manuscript. 
C.G. and A.R. advised the robot design, experiments, and manuscript.
G.Z. advised the manuscript.

\bmhead{Competing interests}
The authors declare that they have no competing interests.

\bmhead{Data availability}
The datasets generated and analyzed during the current study are available within the main text, methods section, or Supplementary Information. Source data and raw experimental log files are publicly deposited in a public GitHub repository (\url{https://github.com/wgu93/AirPulse_Paper_Figs_and_Data}) and will be permanently archived with a persistent DOI via Zenodo prior to final publication.

\bmhead{Code availability}
The post-processing, analysis scripts, and implementation codes including RLS estimation, PID control, and STAR phase-domain modulation used to support the findings of this study are publicly available in a public GitHub repository (\url{https://github.com/wgu93/AirPulse_Paper_Figs_and_Data}), and will be permanently archived with a persistent DOI via Zenodo prior to final publication. Mathematical formulations and algorithmic details are further documented within the Methods section and Supplementary Information.

% \section*{Declarations}

% Some journals require declarations to be submitted in a standardised format. Please check the Instructions for Authors of the journal to which you are submitting to see if you need to complete this section. If yes, your manuscript must contain the following sections under the heading `Declarations':

% \begin{itemize}
% \item Conflict of interest/Competing interests (check journal-specific guidelines for which heading to use)
% \item Ethics approval and consent to participate
% \item Consent for publication
% \item Materials availability
% \end{itemize}

% \noindent
% If any of the sections are not relevant to your manuscript, please include the heading and write `Not applicable' for that section. 

% \bigskip

% Springer journals and proceedings: \url{https://www.springer.com/gp/editorial-policies}

% Nature Portfolio journals: \url{https://www.nature.com/nature-research/editorial-policies}

% \textit{Scientific Reports}: \url{https://www.nature.com/srep/journal-policies/editorial-policies}

% BMC journals: \url{https://www.biomedcentral.com/getpublished/editorial-policies}
% \end{flushleft}

\begin{appendices}
\section{Mass and inertial properties}\label{secA1}

\begin{table}[htbp]
\centering
\tiny % Tiny font size allows all 11 columns to fit side-by-side cleanly
\setlength{\tabcolsep}{1.8pt} % Very narrow spacing to prevent text overflow
\caption{\rev{\textbf{Variation of Center of Gravity (CG) and Moments of Inertia with Flap Angle.} Data corresponding to the structural model of the AirPulse robot across a single half-stroke (flap angles from $+70^\circ$ to $-70^\circ$). $I_{xx}$, $I_{yy}$, and $I_{zz}$ denote the body-frame inertias, highlighting the dynamic structural coupling.}}
\label{tab:inertia_variation}
\begin{tabular}{S[table-format=-2.0] S[table-format=2.2] S[table-format=-2.3] S[table-format=-1.3] S[table-format=-2.3] S[table-format=6.3] S[table-format=-1.3] S[table-format=-5.3] S[table-format=6.3] S[table-format=-2.3] S[table-format=6.3]}
\toprule
{\textbf{Flap}} & {\textbf{Mass}} & \multicolumn{3}{c}{\textbf{CG Position (mm)}} & \multicolumn{6}{c}{\textbf{Inertia Tensor Components ($\mathbf{g\cdot mm^2}$)}} \\
\cmidrule(lr){3-5} \cmidrule(l){6-11} 
{\textbf{Angle ($^\circ$)}} & {\textbf{(g)}} & {$\mathbf{X}$} & {$\mathbf{Y}$} & {$\mathbf{Z}$} & {$\mathbf{I_{xx}}$} & {$\mathbf{I_{xy}}$} & {$\mathbf{I_{xz}}$} & {$\mathbf{I_{yy}}$} & {$\mathbf{I_{yz}}$} & {$\mathbf{I_{zz}}$} \\ 
\midrule
70  & 25.7 & -19.019 & -0.036 & 15.738  & 137531.489 & -4.664 & -2575.648  & 138174.711 & 25.891  & 84482.821  \\
60  & 25.7 & -17.179 & -0.036 & 14.005  & 143507.811 & -3.452 & 3476.369   & 125061.027 & 23.898  & 103996.565 \\
50  & 25.7 & -15.493 & -0.036 & 11.542  & 148829.781 & -2.374 & 7739.866   & 109302.177 & 21.326  & 127007.636 \\
40  & 25.7 & -14.014 & -0.036 & 8.424   & 153383.099 & -1.448 & 9825.859   & 92596.732  & 18.267  & 151114.066 \\
30  & 25.7 & -12.787 & -0.036 & 4.746   & 157074.168 & -0.686 & 9641.813   & 76770.927  & 14.813  & 173736.031 \\
20  & 25.7 & -11.848 & -0.036 & 0.619   & 159827.533 & -0.097 & 7397.975   & 63564.126  & 11.059  & 192417.066 \\
10  & 25.7 & -11.226 & -0.036 & -3.831  & 161583.752 & 0.314  & 3576.300   & 54424.174  & 7.095   & 205112.073 \\
0   & 25.7 & -10.940 & -0.036 & -8.469  & 162298.280 & 0.542  & -1134.103  & 50337.176  & 3.010   & 210427.677 \\
-10 & 25.7 & -10.999 & -0.036 & -13.154 & 161941.758 & 0.584  & -5927.018  & 51712.092  & -1.108  & 207786.092 \\
-20 & 25.7 & -11.401 & -0.036 & -17.743 & 160501.829 & 0.436  & -9983.595  & 58333.963  & -5.171  & 197492.888 \\
-30 & 25.7 & -12.133 & -0.036 & -22.098 & 157986.310 & 0.095  & -12578.437 & 69391.330  & -9.088  & 180700.536 \\
-40 & 25.7 & -13.174 & -0.036 & -26.086 & 154427.288 & -0.440 & -13174.261 & 83574.495  & -12.765 & 159272.098 \\
-50 & 25.7 & -14.492 & -0.036 & -29.585 & 149885.495 & -1.169 & -11493.521 & 99232.765  & -16.103 & 135561.458 \\
-60 & 25.7 & -16.046 & -0.036 & -32.490 & 144454.179 & -2.083 & -7558.135  & 114571.721 & -19.002 & 112136.410 \\
-70 & 25.7 & -17.790 & -0.036 & -34.712 & 138261.672 & -3.171 & -1692.332  & 127866.792 & -21.362 & 91477.824  \\
\bottomrule
\end{tabular}
\end{table}

\section{Forewing contour parameterization}\label{secA1}

\begin{figure}[H]
	\centering
	\includegraphics[width=\textwidth]{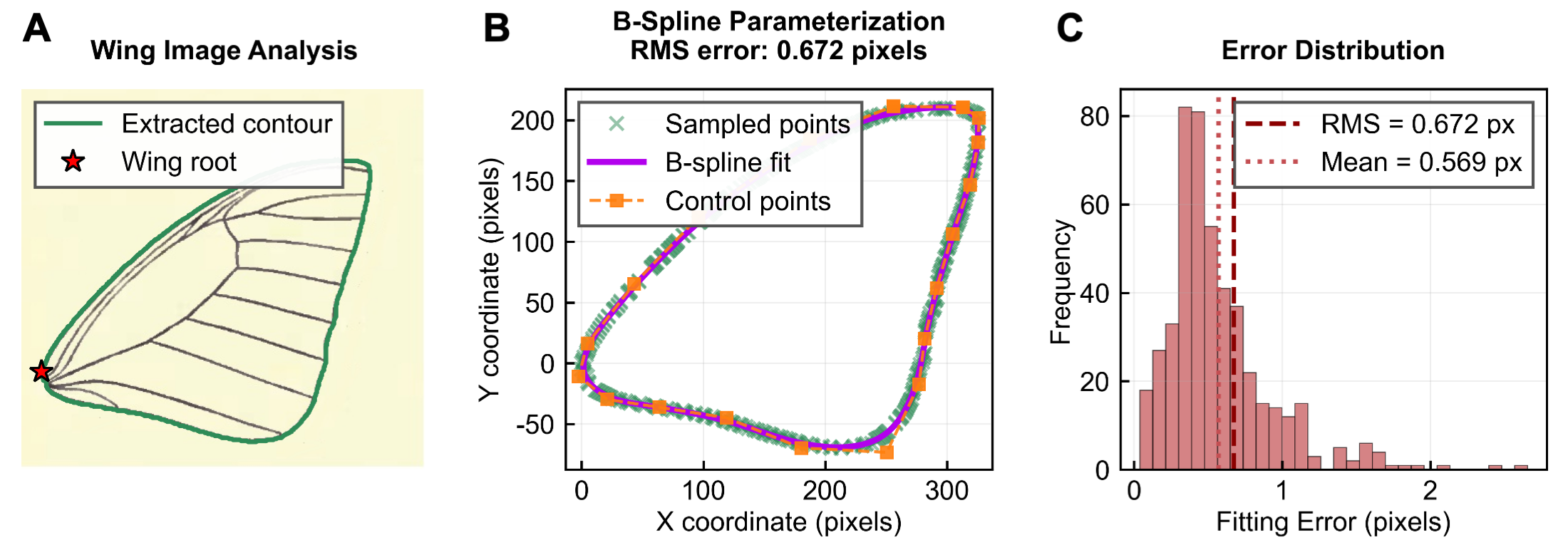}
	\caption{\textbf{Forewing contour parameterization.}}
	\label{fig:SM_WingContourParameterization} 
\end{figure}

\begin{table}[h]
    \centering
    \small % Standard elegant size for simple 3-column data tables
    \caption{\textbf{Parameterization settings and results for forewing contour.}}
    \label{tab:parameterization_results}
    \begin{tabular}{lcc}
        \toprule
        \textbf{Parameter} & \textbf{Value} & \textbf{Units} \\
        \midrule
        Sampled points ($N$) & 477 & points \\
        Control points ($n$) & 22 & points \\
        Data reduction & 95.4\% & -- \\
        RMS fitting error & 0.672 & pixels \\
        Smoothing factor ($\alpha_\mathrm{s}$) & 0.5 & -- \\
        Resampled points ($M$) & 1000 & points \\
        \bottomrule
    \end{tabular}
\end{table}

\begin{table}[h]
    \centering
    \small % Keeps dense 6-column coordinates readable and compact
    \caption{\textbf{Control points for periodic cubic B-spline representation of forewing contour.}}
    \label{tab:control_points}
    \begin{tabular}{c S[table-format=3.6] S[table-format=-3.6] c S[table-format=3.6] S[table-format=-3.6]}
        \toprule
        Index $i$ & {\textbf{$x_i$ (pixels)}} & {\textbf{$y_i$ (pixels)}} & Index $i$ & {\textbf{$x_i$ (pixels)}} & {\textbf{$y_i$ (pixels)}} \\
        \midrule
        0  & 312.908971 & -210.627073 & 11 & 176.467499 &   51.938866 \\
        1  & 256.030813 & -211.484444 & 12 & 220.083362 &   52.401279 \\
        2  & 187.390279 & -184.698024 & 13 & 263.699226 &   52.863692 \\
        3  &  95.897335 & -120.190052 & 14 & 307.315090 &   53.326105 \\
        4  &  43.167102 &  -65.320731 & 15 & 338.611816 &   41.334592 \\
        5  &   4.851825 &  -16.166765 & 16 & 349.589403 &   11.000153 \\
        6  &  -2.349460 &   10.755896 & 17 & 340.247852 &  -29.334286 \\
        7  &  21.089190 &   29.768771 & 18 & 310.587162 &  -69.668725 \\
        8  &  63.968615 &   36.324384 & 19 & 260.607334 & -110.003164 \\
        9  & 118.976443 &   44.915320 & 20 & 190.308367 & -150.337603 \\
        10 & 147.655635 &   48.477453 & 21 &  99.690262 & -190.672042 \\
        \bottomrule
    \end{tabular}
\end{table}

\section{Biological wing venation analysis}

\begin{figure}[H]
	\centering
	\includegraphics[width=\textwidth]{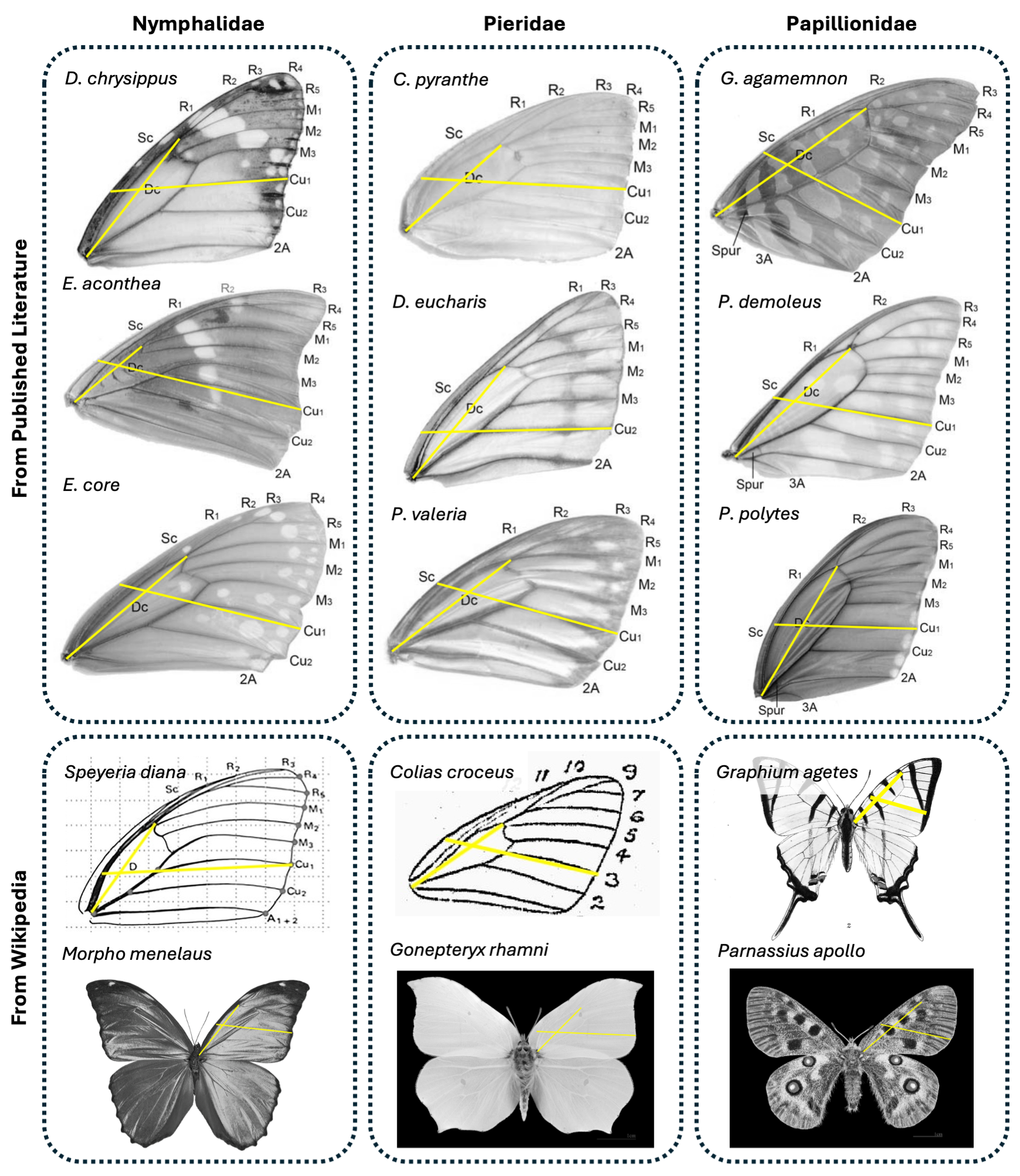} 
	\caption{\textbf{Quantification of forewing Dc-Cu1 venation angle across butterfly families, \rev{showing a representative subset of 15 specimens from the full dataset (\(n = 24\))}.}}
	\label{fig:SM_WingStudies} 
\end{figure}

\section{\rev{Selection of the modulation function and phase-drift compensation}}
\label{app:star}

We considered two candidate modulation functions for STAR with a common sign convention: (i) $p=0.5-A(t)\sin\omega$, and (ii) $p=0.5-A(t)\cos\omega$. The cosine form is preferred because, with the phase origin at mid-downstroke, it eliminates modulation-dependent integration offsets associated with the initial condition. For the cosine variant, \begin{align} \frac{d\omega}{dt} &=\frac{\pi f}{0.5-A(t)\cos\omega}, \label{eq:standard_s3m_phase}\\ \left[0.5-A(t)\cos\omega\right]\dot{\omega} &=\pi f, \end{align} and therefore \begin{equation} \frac{d}{dt} \left[ 0.5\omega-A(t)\sin\omega \right] = \pi f-\dot{A}(t)\sin\omega. \end{equation} Integration gives \begin{equation} 0.5\omega(t)-A(t)\sin\omega(t) = \pi f t -\int_0^t\dot{A}(\tau)\sin\omega(\tau)\,d\tau +C_0. \label{eq:cos_integrated} \end{equation} Under antisymmetric modulation, $A_L(t)=-A_R(t)=A(t)$, subtraction of the corresponding left- and right-wing equations yields \begin{equation} \begin{aligned} \Delta(t) =2\Bigg\{& A(t)\left[\sin\omega_L(t)+\sin\omega_R(t)\right]\\ &-\int_0^t\dot{A}(\tau) \left[\sin\omega_L(\tau)+\sin\omega_R(\tau)\right]\,d\tau +C_{L0}-C_{R0} \Bigg\}, \end{aligned} \label{eq:delta_cos} \end{equation} where $\Delta=\omega_L-\omega_R$ is the phase difference. For the sine candidate $p=0.5-A(t)\sin\omega$, an analogous derivation gives \begin{equation} \begin{aligned} \Delta(t) =2\Bigg\{& -A(t)\left[\cos\omega_L(t)+\cos\omega_R(t)\right]\\ &+\int_0^t\dot{A}(\tau) \left[\cos\omega_L(\tau)+\cos\omega_R(\tau)\right]\,d\tau +C_{L0}-C_{R0} \Bigg\}. \end{aligned} \label{eq:delta_sin} \end{equation} The difference becomes apparent from the integration constants. For symmetric initialization at mid-downstroke, $\omega_L(0)=\omega_R(0)=0$, Eq.~\eqref{eq:cos_integrated} gives $C_{L0}=C_{R0}=0$, independently of $A(0)$. In contrast, for the sine variant, \begin{equation} C_{L0}=A(0), \qquad C_{R0}=-A(0), \qquad C_{L0}-C_{R0}=2A(0), \end{equation} such that the integrated phase dynamics retain an $A(0)$-dependent offset. The cosine form therefore removes this modulation-dependent initial-condition bias. Changes in $A(t)$ can nevertheless introduce an additional history-dependent term through $\dot{A}$. This transient term can be eliminated by augmenting the cosine phase dynamics as \begin{equation} \label{eq:extended_s3m_phase} \frac{d\omega}{dt} = \frac{\pi f+\dot{A}(t)\sin\omega} {0.5-A(t)\cos\omega}. \end{equation} Indeed, substitution into \begin{equation} \frac{d}{dt} \left[0.5\omega-A(t)\sin\omega\right] = \left[0.5-A(t)\cos\omega\right]\dot{\omega} -\dot{A}(t)\sin\omega \end{equation} gives exactly \begin{equation} \frac{d}{dt} \left[0.5\omega-A(t)\sin\omega\right] =\pi f, \end{equation} thereby removing the history-dependent $\dot{A}$ term during modulation transients. For constant $A$, Eq.~\eqref{eq:extended_s3m_phase} reduces exactly to the standard STAR formulation in Eq.~\eqref{eq:standard_s3m_phase}.

\section{Experimental force--torque mapping from flapping kinematics}

\begin{figure}[H]
	\centering
	\includegraphics[width=0.7\textwidth]{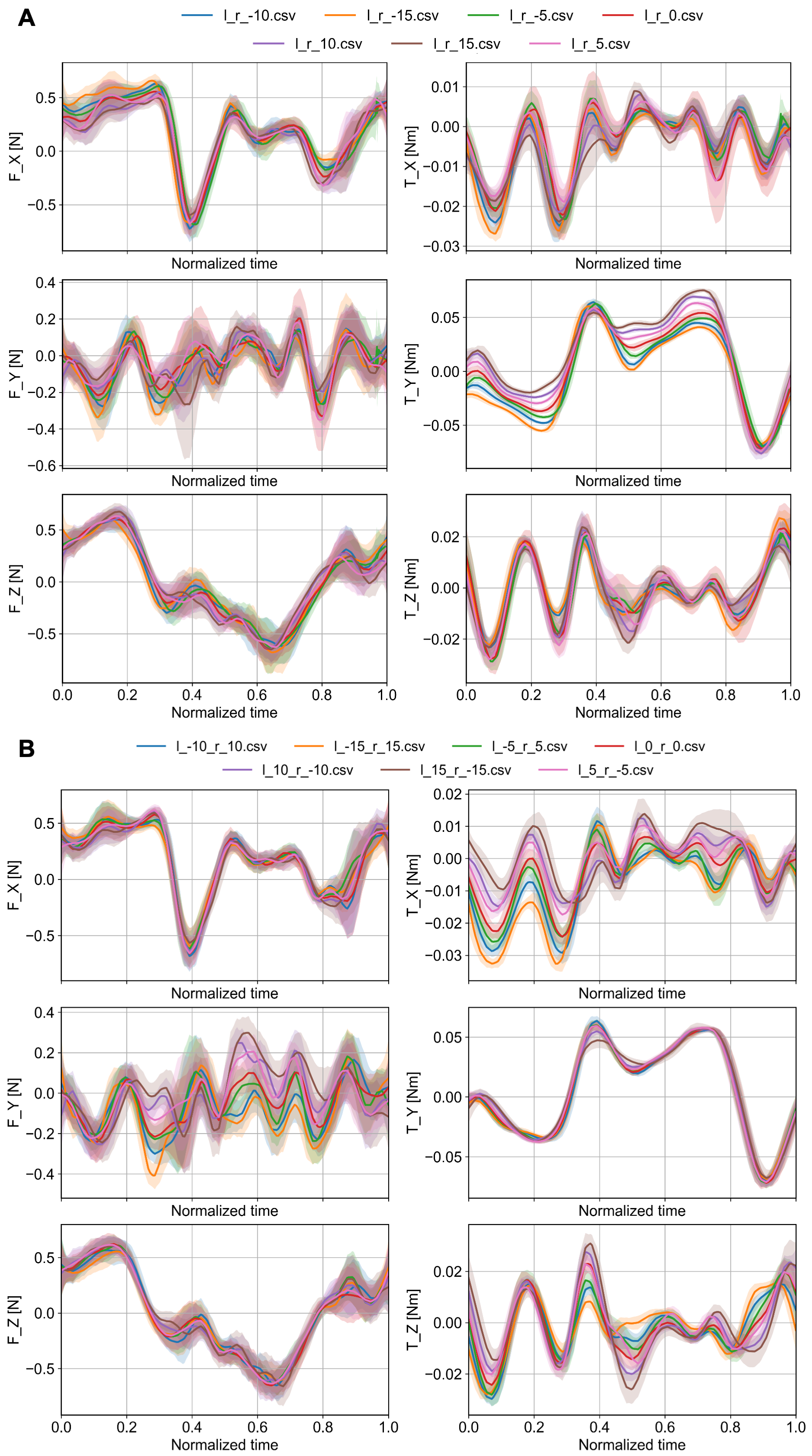} 
	\caption{\textbf{Forces and torques generated by angle offset modulation.} (\textbf{A}) Symmetric modulation (\texttt{l\_r\_X}): both wings have an angle offset $X$ (in deg, positive means upward). 
(\textbf{B}) Antisymmetric modulation (\texttt{l\_X\_r\_Y}): left and right wings have offsets $X$ and $Y$, respectively. 
Colors indicate different modulation settings.}
	\label{fig:SM_AngleOffsetVsForceTorque}
\end{figure}

\begin{figure}[H]
    \centering
    \includegraphics[width=0.85\linewidth]{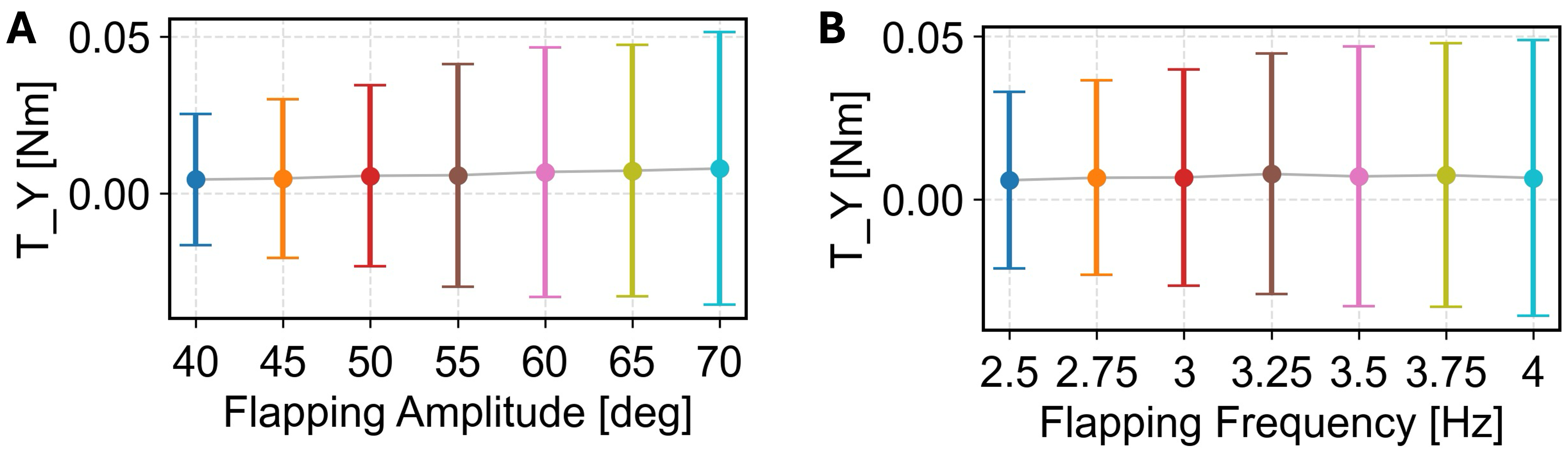}
    \caption{\rev{\textbf{Effects of symmetric flapping-amplitude and frequency modulation on cycle-averaged pitch moment.} (\textbf{A}) Amplitude modulation and (\textbf{B}) frequency modulation applied identically to the two wings. The cycle-averaged pitch moment remains nearly unchanged across the tested settings and varies less than under the selected stroke-offset and STAR pitch channels shown in the main text. Under ideal bilateral symmetry, roll- and yaw-moment contributions cancel, although residual moments may occur due to wing and actuator asymmetries. Colors follow Fig.~\ref{fig:Fig4_FT}.}}
    \label{fig:SM_CycleAveragedPitchMoment}
\end{figure}
% For amplitude modulation, the plotted cycle-averaged T Y  values are approximately: Φ [deg] T Y   [mNm]  40 4.6  45 4.9  50 5.6  55 6.0  60 7.0  65 7.3  70 8.1   Thus, Δ T Y   amp  ≈8.1−4.6=3.5 mNm  over 40 ∘ –70 ∘ . For frequency modulation, I obtain approximately: f [Hz] T Y  [mNm]  2.5 4.3 2.75 5.2  3.0 5.2  3.25 6.4  3.5 5.5  3.75 6.0  4.0 4.9   giving Δ T Y   f  ≈6.4−4.3=2.1 mNm. 
% | Modulation              | Peak-to-peak cycle-averaged pitch-moment variation |
% | ----------------------- | -------------------------------------------------: |
% | Amplitude               |                                       **3.5 mN·m** |
% | Frequency               |                                       **2.1 mN·m** |
% | Symmetric stroke offset |                                       **~21 mN·m** |
% | Symmetric STAR          |                                       **~15 mN·m** |

\section{State estimation and PID tuning}
\label{app:pid}
Attitude estimation was implemented on the ESP32-S3 using the quaternion-based Madgwick filter~\cite{madgwick2011estimation}. The filter was selected because it can be evaluated at the 100~Hz control rate with low memory and computational demand on the onboard processor.
% Accurate and low-latency attitude estimation is critical for stable flight of the flapping-wing platform. We implemented a quaternion-based Madgwick filter~\cite{madgwick2011estimation} on the ESP32-S3 (dual-core 240~MHz with floating-point unit) to fuse IMU measurements in real time. This approach was selected for its low computational cost, fast convergence, and robustness under dynamic motion, outperforming complementary filters in accuracy and avoiding the high complexity and matrix operations of extended Kalman filters.

The Madgwick filter estimates orientation by minimizing accelerometer and magnetometer errors using a gradient descent approach. For 9-axis estimation, tri-axial magnetometer data provides a reference for yaw, reducing drift and enabling absolute heading in the Earth frame. In contrast, the 6-axis variant uses only accelerometer and gyroscope data, resulting in drift-prone yaw estimates from integrated rates. Since all flights were conducted indoors, we employed the 6-axis Madgwick filter and initialized the yaw to zero after \ac{IMU} startup. Computation involves only basic vector and trigonometric operations, allowing high-frequency updates suitable for the embedded platform.

Attitude estimates from the Madgwick filter serve as inputs to a \ac{PID} controller for yaw stabilization. For pitch control, the measured angle is first processed by the \ac{RLS} algorithm to extract the low-frequency mean before being fed to a \ac{PID} controller. Gains were tuned empirically in flight to balance responsiveness and damping, accounting for inherent body-wing oscillations. The controller outputs differential commands to the \ac{CPG}, which converts them into \ac{PWM} signals driving the left and right wings, enabling stable maneuvers and satisfactory tracking performance. Pitch \ac{PID} gains are $K_p/K_i/K_d = 0.6/0.45/0.05$ for stroke-offset modulation and $0.6/0.7/0.07$ for STAR modulation. For yaw control, $K_p=0.15$ for stroke-offset modulation and $K_p=0.17$ for STAR modulation, with $K_i=K_d=0$.

\begin{figure}[H]
	\centering
	\includegraphics[width=0.9\textwidth]{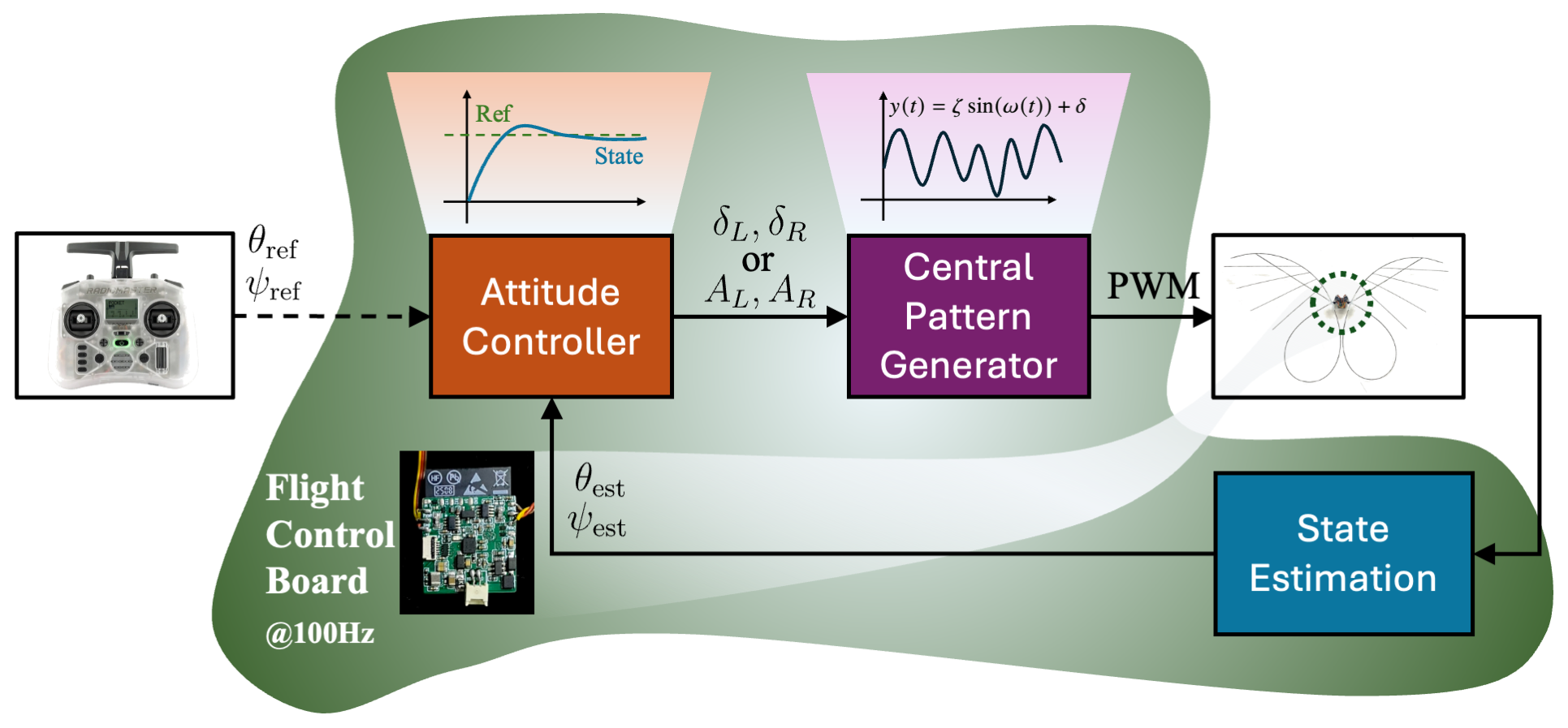} 
	\caption{\textbf{Flight control architecture for attitude tracking.} Raw \ac{IMU} data are fused by the onboard estimator using the Madgwick filter to provide real-time attitude feedback to the controller, which generates modulatory commands based on either wireless or preset references. \ac{CPG} then converts these commands into rhythmic \ac{PWM} outputs that actuate the left and right wing servos independently. All computations are executed on the flight control board at 100~Hz.}
	\label{fig:SM_ControllerArchitecture} 
\end{figure}

% \section{Supplementary movie}

% \paragraph{Caption for Movie S1.}
% \textbf{Overview of the AirPulse robot and demonstration of autonomous flight.}

% \paragraph{Caption for Movie S2.}
% \textbf{\rev{Forewing--hindwing phase lag and passive feathering in the compliant wing design.}} 

% \paragraph{Caption for Movie S3.}
% \textbf{\rev{Qualitative comparison of wingbeat-synchronous body
% motion in a butterfly and the AirPulse robot.}} 

% \paragraph{Caption for Movie S4.}
% \textbf{Demonstration of a climbing maneuver.}

% \paragraph{Caption for Movie S5.}
% \textbf{Demonstration of a turning maneuver.}

\end{appendices}

\bibliography{sn-bibliography}% common bib file

\end{document}